\documentclass[preprint,11pt,3p,authoryear,nopreprintline]{elsarticle}

\usepackage{amssymb}
\usepackage{amsthm}

\usepackage{amsmath,amssymb,graphicx}
\usepackage{algorithm} 
\usepackage{algorithmic}

\usepackage{cases}
\usepackage[caption=false]{subfig}
\usepackage{caption} 
\usepackage{hyperref}
\hypersetup{
	colorlinks=true,
	linkcolor=blue,
	filecolor=magenta,      
	urlcolor=cyan,
}
\journal{European Journal of Operational Research}
\usepackage{booktabs}
\usepackage{graphicx}
\newtheorem{definition}{Definition}[section]
\usepackage{setspace}
\usepackage{tabularx}
\newcolumntype{C}{>{\centering\arraybackslash}X}

\begin{document}

\begin{frontmatter}



\title{An Expandable Machine Learning-Optimization Framework to Sequential Decision-Making}


\author[label1]{Dogacan Yilmaz}   
\ead{dy234@njit.edu}
\author[label2]{\.I. Esra B\"uy\"uktahtak{\i}n\corref{cor1}}
\ead{esratoy@vt.edu}

\address[label1]{Department of Mechanical and Industrial Engineering, New Jersey Institute of Technology, Newark, NJ 07102, USA}  
\address[label2]{Department of Industrial and Systems Engineering, Virginia Tech, Blacksburg, VA 24061, USA}

\cortext[cor1]{Corresponding author}

\nonumnote{The final version of this article has been accepted for publication in European Journal of Operational Research and can be accessed via \href{https://doi.org/10.1016/j.ejor.2023.10.045}{https://doi.org/10.1016/j.ejor.2023.10.045}.}

\begin{abstract}
We present an integrated prediction-optimization (PredOpt) framework to efficiently solve sequential decision-making problems by predicting the values of binary decision variables in an optimal solution. We address the key issues of sequential dependence, infeasibility, and generalization in machine learning (ML) to make predictions for optimal solutions to combinatorial problems. The sequential nature of the combinatorial optimization problems considered is captured with recurrent neural networks and a sliding-attention window. We integrate an attention-based encoder-decoder neural network architecture with an infeasibility-elimination and generalization framework to learn high-quality feasible solutions to time-dependent optimization problems. In this framework, the required level of predictions is optimized to eliminate the infeasibility of the ML predictions. These predictions are then fixed in mixed-integer programming (MIP) problems to solve them quickly with the aid of a commercial solver. We demonstrate our approach to tackling the two well-known dynamic NP-Hard optimization problems: multi-item capacitated lot-sizing (MCLSP) and multi-dimensional knapsack (MSMK). Our results show that models trained on shorter and smaller-dimensional instances can be successfully used to predict longer and larger-dimensional problems. The solution time can be reduced by three orders of magnitude with an average optimality gap below 0.1\%. We compare PredOpt with various specially designed heuristics and show that our framework outperforms them. PredOpt can be advantageous for solving dynamic MIP problems that need to be solved instantly and repetitively.
\end{abstract}


\end{frontmatter}



\section{Introduction}

The goal of this paper is to contribute to bridging the gap between two traditionally distinct research areas, Operations Research (OR) and Machine Learning (ML), to solve NP-hard sequential decision-making problems. OR is a discipline that aims to find the best decisions for complex problems through mathematical modeling and optimization, while ML focuses on learning from the data without explicitly programming it. In this paper, we tackle a very hard category of OR problems known as combinatorial optimization problems by innovatively combining a machine translation learning framework with an optimization-learning framework. 

Our objective is to substantially reduce the solution times of such combinatorial problems while providing high-quality feasible solutions, which could be very useful in practical applications. In most industrial settings, such as finance, health, energy, and manufacturing systems, OR problems with the same structures are repeatedly solved with different parameters. For such settings, a reduced solution time can provide an immense advantage to decision-makers. In this study, we present an expandable framework based on a sequence-to-sequence neural machine translation system to solve sequentially dependent optimization problems with all feasible predictions, which are either optimal or very close to optimal.

\citet{yilmaz2023learning} present one of the pioneering studies that utilize an ML-based prediction methodology to reduce the solution times of repeatedly-solved combinatorial problems in a multi-period setting. Specifically, \citet{yilmaz2023learning} harness bidirectional Long Short-Term Memory (LSTM) networks to predict binary decision variables that denote the production decision in the capacitated lot-sizing problem. Models are trained using the solutions of problems that are solved to optimality using CPLEX. Then, predictions are generated using the trained model for new instances and added to the problem as constraints to be resolved using CPLEX. Their results show that, depending on the hardness of the problem, the solution time can be decreased by up to six orders of magnitude without a significant increase in the infeasibility or optimality gap, especially when problems with longer planning horizons are considered. In \citet{yilmaz2023learning}, predictions for problems with longer planning horizons are generated by concatenating the predictions generated by models trained using problems with shorter planning horizons; e.g., predictions for 270-period problems are generated as three independent prediction sets first from 1 to 90, second from 91 to 180, and third from 181 to 270 using the model that trained 90 periods with. However, this approach disregards the dependence between consecutive sets; e.g., prediction sets 1 to 90 and 91 to 180 are generated independently, which may impact the quality of the predictions. Also, it does not provide a methodology to accommodate problems with periods that are not exact multiples of the training periods. In this study, we integrate the attention-based encoder-decoder network presented in \citet{luong2015effective} into a prediction-optimization framework to overcome the sequential dependence problems, since the output sequence is allowed to be of variable size. 

Furthermore, in \citet{yilmaz2023learning}, the authors show that the percentage of predicted binary variables that are added to the problem can have a major impact on the quality of the solutions. In other words, the use of high-level predictions may lead to high infeasibility, since the predicted values are fixed in the problem. \citet{yilmaz2023learning} states that fixing all predictions of binary variables may cause infeasibility in one of the two problems in some cases, which is highly undesirable. We propose an iterative algorithm to eliminate infeasible predictions to overcome this issue. Another important research question has been the generalizability of such ML approaches in predicting optimal solutions. For example, could a smaller-dimensional model with a few items be used to predict a higher-dimensional problem with a larger set of items? In this paper, we develop new algorithms to address those main research challenges in utilizing ML approaches to predict optimal solutions to NP-Hard problems.

In practice, combinatorial optimization problems are frequently solved with changing input data. In this study, we focus on solving two core NP-Hard problems where decisions are made over a time horizon: Multi-item Capacitated Lot Sizing Problem (MCLSP) and Multi-stage Knapsack Problem (MSMK). In MCLSP, sequentially-dependent decisions are made to determine the production and inventory levels for the planning horizon, considering the changing costs, demand, and capacity. The lot-sizing is one of the most important and difficult problems in production planning. Its variants have a wide range of applications in numerous fields, including the production, medical, and chemical industries \citep{karimi2003capacitated}. Another example of dynamic combinatorial optimization problems is when the stability over the solution can possess significant importance since it might incur setup costs frequently, i.e., turning on and off an electricity plant can be costly. In such settings, maintaining the stability of the current solution for successive periods is desired. Examples of such problems include MSMK, the core resource allocation problem with stability constraints \citep{bampis2019multistage}, the multistage facility location problem \citep{eisenstat2014facility}, and multistage prize collecting traveling salesperson \citep{bampis2019lpbased}. In this study, we demonstrate our framework through knapsack and lot-sizing, since both are considered as general multi-stage problems and are commonly used to test improvements in the solution algorithms for multi-stage problems \citep{guan2009cutting,buyuktahtakin2023scenario,buyuktahtakin2022stage}.

Our main contribution is to develop an extendable prediction-optimization (PredOpt) framework for sequential decision-making problems to address the key issues of sequential dependence, infeasibility, and generalization in ML prediction for OR. We have innovatively adapted a local attention-based encoder-decoder network, a deep learning tool originally developed for neural machine translation, to learn the optimal solutions for sequentially dependent optimization problems. The sequential nature of the combinatorial optimization problems considered is captured with recurrent neural networks and a sliding-attention window. The models can be trained on short-period problems to learn long-period problems. Furthermore, we present a specific prediction algorithm that enables trained models with smaller items to be generalized to predict instances with much larger items. Additionally, we develop an iterative algorithm to quickly verify the feasibility of the predicted solutions and determine the optimal level of predictions. The resulting predictions practically eliminate the possibility of infeasible solutions resulting from the predictions. We demonstrate our general framework to tackle MCLSP and MSMK without assuming any specific details of the problem and, therefore, without tailoring it to a specific problem. We show the computational efficiency of the PredOpt framework and the quality of the predictions in terms of the solution time reduction and optimality gap metrics.

\vspace{-0.3cm}
\section{Literature Review and Contributions}

The use of ML for OR has recently gained much interest with successful ML applications in different problems. The closest study to ours is \citet{frejinger2019language}. They use a similar approach in which an attention-based encoder-decoder neural network architecture of \citet{bahdanau2014neural} is used to predict fast solutions to a combinatorial optimization problem under imperfect information. They demonstrate creating input and output vocabularies to represent the optimization problem as a pair of input and output languages. Our study differs significantly from \citet{frejinger2019language} in several key points. First, our focus is on predicting binary variables solely for use in the Mixed-Integer Linear Programming (MILP) solver to reduce the solution time of the problem. We achieve this objective with a methodology to determine the highest level of prediction that will not cause infeasible predictions. On the other hand, \citet{frejinger2019language} have predicted only tactical decisions since fully detailed solutions are not needed at the time of prediction. Therefore, the type and usage of the decisions made by the decision variables are different. Although \citet{frejinger2019language} present results on generalizability to different instances, we specifically focus on training using shorter and smaller-dimension instances to predict longer and higher-dimension, therefore, harder problems. We also propose a specially designed item-wise expansion algorithm for generalization and present detailed computational results to demonstrate the generalization capability of our PredOpt framework. Furthermore, \citet{frejinger2019language} utilizes the neural machine translation model presented by \citet{bahdanau2014neural}, whereas we employ the architecture presented in \citet{luong2015effective} with local attention to capture sequential dependencies. In another study, \citet{larsen2021predicting} focus on solving the same optimization problem as \citet{frejinger2019language} with a similar motivation. Rather than a language translation system, they use multilayer perceptrons to predict tactical descriptions of operational solutions in a less detailed aggregation compared to \citet{frejinger2019language}. In a recent study, \citet{bertsimas2021voice} used optimal classification trees to learn about the solution strategies of optimization problems. \citet{bertsimas2021online} build upon that framework to solve parametric mixed-integer quadratic optimization problems without requiring a solver.

\citet{anderson2021generative} present a generative neural network design to reduce the resolution times of repetitively solved optimization problems. The presented framework contains two neural networks: A generator to predict the values of binary decision variables and a discriminator to predict the value of the objective function when those variables are fixed. Although their motivation is similar to ours, the methodology they developed is quite different from ours. They utilize a model based on generative adversarial networks, whereas we employ an encoder-decoder network. While both studies predict the values of binary decision variables, their usages are dissimilar. \citet{anderson2021generative} feed predictions to a discriminator to obtain a prediction of the objective when those variables are fixed in the problem and then use them as a warm start for the optimizer. We utilize predictions of binary variables in a feasibility check loop to determine the optimal prediction level and solve the problem with fixed variables at a determined level. They show their framework for the transient gas optimization problem, and they are able to reduce the solution time by 60.5\%. We demonstrate our prediction framework to solve MCLSP and MSMK.

\citet{zamzam2020learning} utilize neural networks to learn the optimal solutions of an AC optimal power flow problem. They also emphasize the feasibility of the generated predictions. In \citet{pan2019deepopf}, a neural network-based framework is presented to handle the infeasibility of the predictions when solving the optimal power flow problem. They ensure feasibility by adjusting the limits of the constraints during training. Another methodology to eliminate infeasibility is presented in \citet{donti2021dc3} by enforcing constraints during training for the AC optimal power flow problem. 



\citet{vinyals2015pointer} present pointer networks to solve combinatorial optimization problems such as the travel salesperson and the convex planar hull problem. The encoder-decoder architecture uses an attention mechanism as a pointer to select an input element as output at each decoding time step. The proposed architecture is suitable for combinatorial optimization problems where the size of the output depends on the size of the input sequence. Even though their seminal work shares the same motivation as ours, our problem of interest is to have a time-wise extendable framework with a constant-sized output rather than a mapping to the input elements to handle variable-sized output when decoding.

In a recent study, \citet{bello2016neural} present a methodology for using reinforcement learning as an alternative to the pointer networks presented in \citet{vinyals2015pointer}, where supervised learning is not desired. The proposed framework can achieve solutions close to optimal when solving the traveling salesperson problem. \citet{nazari2018reinforcement} generalize the framework presented in \citet{bello2016neural} to handle more complex problems that system dynamics change over time. The authors show that the learned policy can achieve near-optimal solutions for the capacitated vehicle routing problem. \citet{deudon2018learning} study the solution of the traveling salesperson problem by replacing the LSTM in \citet{bello2016neural} with an attention mechanism and improving the predicted solution with an established heuristic.

\citet{kool2018attention} present a new model based on attention and a training methodology. They follow a transformer architecture with multi-head attention instead of the encoder-decoder with recurrent neural networks. Their attention-based model focuses on learning heuristics for solving the traveling salesperson and vehicle routing problem and achieving a performance level close to highly-specialized algorithms. \citet{joshi2019efficient} propose a learning-based approach to solve the travel salesperson problem by combining graph convolutional networks and beam search. Their approach is non-autoregressive and can scale better to larger instances. \citet{lu2020learning} present a learn-to-improve framework for solving capacitated vehicle routing problems using deep reinforcement learning. Their ensemble method outperforms the classical OR solution approaches. \citet{li2021learning} introduce a study to solve large-scale vehicle routing problems by identifying and solving smaller subproblems. \citet{bushaj2022simulation} introduce a reinforcement learning approach for the optimization of epidemic control by comparing the decisions of the learners and the actions of the government during COVID-19. In a recent study, \citet{yilmaz2023a} utilize a deep reinforcement learning framework to solve scenario-based two-stage stochastic programs with a multi-agent structure. For a detailed review of the use of ML for OR, we refer to the excellent review by \citet{bengio2021machine}.


\subsection*{Key Contributions of the Study}
While there has been a significant advancement in the use of ML for OR in recent years, as discussed above, an extensive research gap exists in the literature. In particular, frameworks that utilize time-dependent learning models and algorithms to enforce the feasibility and integration of advanced computing capabilities of OR solvers with promising results of ML can be further investigated to solve hard mathematical programs. These research limitations inspire our study. Our motivation is to develop new algorithms that innovatively adapt neural translation deep learning architectures to predict optimal solutions to NP-hard optimization problems. Our goal is to reduce solution times while ensuring feasible and near-optimal solutions when problems with similar structures are repeatedly solved. Our key contributions are listed below:
\begin{enumerate}
\item To our knowledge, this is the first study to explore an encoder-decoder approach to predict optimal solutions to sequential decision-making problems. Specifically, we have designed the PredOpt framework based on an encoder-decoder with an attention mechanism to capture the dynamic relationship between input parameters and optimal solutions to MCLSP and MSMK problems. 
\item Our machine learning approach involves a local attention structure with a time window to better capture the association among the problem periods, as the current prediction period is more closely related to the preceding and subsequent periods than to the entire sequence. This reduces the computational cost compared to the global attention mechanism as in the classical LSTM and enables selectively focusing on a few close-by periods near the decision point, which better suits the sequentially-dependent problems.
\item The presented PredOpt framework with an encoder-decoder mechanism learns from shorter problems to solve much longer ones. Furthermore, we have developed a new generalization algorithm to create predictions abstracted from small problems to be used for larger problems. Specifically, we have shown that trained models are generalizable by training them with a few items to predict problems with a large number of items. This approach results in a significant time gain in training set generation and training time. This is also critical because once a model with a smaller dimension is trained, it can be used to solve numerous problems with larger dimensions.
\item We tackle the problem of infeasibility in ML predictions by proposing a new iterative methodology to find feasible predictions and a favorable prediction level that decreases the solution time. This algorithm includes predicting tight constraints, which in turn is used to create a relaxed problem to quickly check the feasibility. We then utilize a second iteration to ensure that the prediction level determined by the relaxation is updated if necessary.
\item We generate benchmark MCLSP and MSMK instances and compare the computational performance of the PredOpt framework with the state-of-the-art commercial solver CPLEX version 20.1.0 and heuristics in terms of the optimality gap and solution time. Our results show that the solution time can be improved by up to a factor of 7,236 with an optimality gap of only 0.11\% on average, while ensuring that the predictions used to obtain the solutions are feasible.
\item The presented PredOpt framework can be quite beneficial for applications where problems with similar structures are repeatedly solved, which are common in various industries from manufacturing to electronics, energy and healthcare systems, and the public good \citep{finnah2022integrated, buyuktahtakin2018new, yin2021covid, bushaj2022risk}. 
\end{enumerate}

The remainder of the paper is as follows. Section \ref{Problems} presents the formulations for MCLSP and MSMK. Section \ref{Methodology} presents the encoder-decoder model and the PredOpt framework. Section \ref{Implementation} explains the implementation steps, experimentation environment, and metrics used to measure the quality of the PredOpt framework. Section \ref{predoptResults} presents the results obtained using the PredOpt framework. Section \ref{predoptConclusions} presents conclusions with future directions.
 
\section{Problems}
\label{Problems}
\vspace{-0.2cm}
In this section, we present two problem formulations of specific interest in this study: Multi-item Capacitated Lot Sizing Problem (MCLSP) and Multi-stage Knapsack Problem (MSMK).
\vspace{-0.2cm}

\subsection{Multi-item Capacitated Lot Sizing Problem}
\label{mclsp}

MCLSP is an extension of the single-item CLSP in which multiple items compete for a shared capacity at each time period in a production planning setting. MCLSP decides the production and inventory amount for each item in each period by minimizing the sum of costs, which includes production, setup, and inventory costs. The demand, which is known in advance, is satisfied for each item and period pair if possible, and back-ordering is not allowed. MCLSP is NP-Hard \citep{bitran1982computational}, and it has variations that include setup times, pricing decisions, lost sales, shortage costs, safety stocks, and demand uncertainty that are used in the production and manufacturing industries \citep{maes1988multi}.

MCLSP is formulated as a mixed-integer program (MIP), where $T$ is the number of periods in the planning horizon, and $I$ is the number of items considered. For each pair of items $i \in \left\{1,\ldots,I\right\}$ and period $t \in \left\{1,\ldots,T \right\}$, the problem parameters are demand $d_{it}$, unit production cost $p_{it}$, setup cost $f_{it}$, and unit inventory holding cost $h_{it}$. Production capacity $c_{t}$ is the total capacity available for each period $t \in \left\{1,2,\ldots,T \right\}$. All parameters of the MCLSP are assumed to be non-negative. The decision variables $x_{it}$ and $s_{it}$ denote the number of units produced and the ending inventory of item $i$ at period $t$, respectively. The decision to produce is binary $y_{it}$, which is set to 1 if the item $i$ is produced during the period $t$ and 0 otherwise. The MCLSP formulation:

\begin{subequations}
\label{mclsp_formulation}
\begin{eqnarray}
\min &&     \sum_{i=1}^{I} \sum_{t=1}^{T} (p_{it}x_{it}  + f_{it}y_{it} + h_{it}s_{it})\label{mclsp_obj}\\
\textrm{s.t.} &&
      s_{i,t-1} + x_{it} - d_{it} = s_{it} \quad \quad \forall i=1,\ldots,I,\quad \forall t=1,\ldots,T  \label{mclsp_c1}\\
      && \sum_{i=1}^{I} x_{it} \leq c_{t} \quad \quad \forall t=1,\ldots,T  \label{mclsp_c2}\\
      && x_{it} \leq y_{it}c_{t} \quad \quad \forall i=1,\ldots,I,\quad \forall t=1,\ldots,T  \label{mclsp_c3}\\
      && x_{it},s_{it} \geq 0 \quad \quad \forall i=1,\ldots,I,\quad \forall t=1,\ldots,T  \label{mclsp_c4}\\
     && y_{it} \in \{0,1\} \quad \quad \forall i=1,\ldots,I,\quad \forall t=1,\ldots,T.  \label{mclsp_c5}
\end{eqnarray}
\end{subequations}

The sum of production, setup, and holding costs is minimized in the objective function \eqref{mclsp_obj} over each item $i \in \left\{1,\ldots,I\right\}$ and period $t \in \left\{1,\ldots,T \right\}$. The flow of inventory is established with constraints \eqref{mclsp_c1}. Specifically, the demand in period $t$ for item $i$ is fulfilled with the inventory at the end of period $t-1$ and the units produced in period $t$, and the remaining units are set to be the inventory at the end of period $t$. Constraints \eqref{mclsp_c2} ensure that the sum of items produced of all types is limited by capacity for each period $t$. Constraints \eqref{mclsp_c3} assert the related setup cost if the item $i$ is produced in period $t$. Finally, constraints \eqref{mclsp_c4} enforce non-negativity, and constraints \eqref{mclsp_c5} ensure $y_{it}$ are binary. The parameter $s_{i0}$ represents the initial inventory of item $i$ at time zero and is assumed to be zero.

Both single-item and multi-item versions of the lot-sizing problem have been widely studied in the literature. For example, \citet{florian1980deterministic} provide an exact solution approach based on dynamic programming. Another exact solution approach is developed by \citet{barany1984strong}, where valid ($\ell$,S) inequalities are added to the problem using a separation algorithm. In recent years, inequalities based on dynamic programming and partial-objective inequalities were proposed to solve the multi-item capacitated lot-sizing problem \citep{hartman2010dynamic, buyuktahtakin2018partial}. We refer to the excellent review by \cite{pochet2006production} for exact and heuristic solution methodologies and discussion on the different versions and modifications of the lot-sizing problem. In this study, we utilize the valid ($\ell$,S) inequalities presented by \citet{barany1984strong} with a strategy presented by \citet{buyuktahtakin2018partial} to show that our PredOpt framework performs well even compared to hand-crafted special solution algorithms. Also, relax-and-fix heuristics are commonly used to solve MCLSP and its variations \citep{helber2010fix,toledo2015relax,absi2019worst,pochet2006production}. \citet{absi2019worst} present the famous relax-and-fix heuristic for MCLSP, which we adopt to compare with our framework. Furthermore, \citet{almeder2010hybrid} introduces a heuristic approach that combines ant colony optimization, which is a metaheuristic, and mathematical solvers. \citet{absi2013heuristics} develop a heuristic based on Lagrangian relaxation coupled with an adaptive large neighborhood search algorithm for solving MCLSP with lost sales. Lot-sizing has been studied extensively to include uncertain demand \citep{brandimarte2006multi}, lead time \citep{helber2010fix}, biofuel production \citep{kantas2015multi}, and even carbon emissions \citep{benjaafar2012carbon}. For a detailed review of industrial lot-sizing applications, we refer to \citet{jans2008modeling}. For a detailed review of different lot-sizing variations and solution approaches, we refer to \citet{buschkuhl2010dynamic}.

\subsection{Multi-stage Multi-dimensional Knapsack Problem}
\label{msmk}

The multi-stage multi-dimensional knapsack (MSMK) problem is a dynamic version of the classical knapsack problem where the profit and constraints vary over time. In the multi-stage multi-dimensional knapsack problem, the aim is to find stable solutions over the planning horizon that maximizes profit and satisfies the capacity constraints. In \citet{bampis2019multistage}, the authors state that even the multi-stage single-dimensional knapsack is strongly NP-Hard when $T$ is not fixed. Potential applications involve energy production planning and data center operations where problem parameters, such as prices, energy, raw materials, and resources, change frequently.

MSMK can be formulated as an integer program (IP) where the number of periods considered is denoted by $T$, and the number of items available for the knapsack is denoted by $I$. For each period $t \in \left\{1,\ldots,T \right\}$, binary variable $x_{it}$ takes the value 1 if item $i \in \left\{1,\ldots,I\right\}$ is added to the knapsack and takes the value 0 otherwise. To maintain the stability of the solution at each time stage, a binary variable $y_{it}$ is defined for each item $i \in \left\{1,\ldots,I\right\}$ and period $t \in \left\{1,\ldots,T-1 \right\}$. The variable $y_{it}$ takes value 1 if items $i$'s decision was unchanged from period $t$ to $t+1$, i.e. $x_{it}, x_{i,t+1}=0$, or $x_{it}, x_{i,t+1}=1$, otherwise it takes value 0 if the decision is changed from period $t$ to $t+1$, i.e. $x_{it}=0,  x_{i,t+1}=1$ or $x_{it}=1,  x_{i,t+1}=0$. The profit of each item $i$ in period $t$ is indicated by $p_{it}$, and the stability bonus is indicated by $b_{it}$. The number of available knapsack constraints is denoted by $J$, and the capacity is set to be $c_{jt}$ for each resource constraint $j \in \left\{1,\ldots,J \right\}$, and period $t \in \left\{1,\ldots,T \right\}$. The weight is denoted by $w_{ijt}$ for each item $i \in \left\{1,\ldots,I\right\}$, resource constraint $j \in \left\{1,\ldots,J \right\}$, and period $t \in \left\{1,\ldots,T \right\}$. We adapt the formulation in \citet{bampis2019multistage} to include multiple resource constraints and formulate the MSMK as:

\begin{subequations}
\label{msmk_formulation}
\begin{eqnarray}
\max &&     \sum_{i=1}^{I} \sum_{t=1}^{T} p_{it}x_{it}  + \sum_{i=1}^{I} \sum_{t=1}^{T-1} b_{it}y_{it}\label{msmk_obj}\\
\textrm{s.t.} &&
      \sum_{i=1}^{I} w_{ijt}x_{it} \leq  c_{jt} \quad \quad \forall t=1,\ldots,T,\quad \forall j=1,\ldots,J \label{msmk_c1}\\
      && y_{it} \leq -x_{i,t+1} + x_{it} +1 \quad \quad \forall i=1,\ldots,I,\quad \forall t=1,\ldots,T-1  \label{msmk_c2}\\
      && y_{it} \leq x_{i,t+1} - x_{it} +1 \quad \quad \forall i=1,\ldots,I,\quad \forall t=1,\ldots,T-1  \label{msmk_c3}\\
     && x_{it} \in \{0,1\} \quad \quad \forall i=1,\ldots,I,\quad \forall t=1,\ldots,T \label{msmk_c4}\\
     && y_{it} \in \{0,1\} \quad \quad \forall i=1,\ldots,I,\quad \forall t=1,\ldots,T-1.  \label{msmk_c5}
\end{eqnarray}
\end{subequations}

The sum of profit and stability bonus is maximized in the objective function \eqref{msmk_obj}. Constraints \eqref{msmk_c1} ensure that for each knapsack $j$ and time period $t$, the total weight of the selected items is less than the capacity $c_{jt}$. The constraints \eqref{msmk_c2} and \eqref{msmk_c3} determine the association between the variables $x_{it}$, $x_{i,t+1}$, and $y_{it}$. Specifically, these constraints are linear relaxations of $y_{it} =1- \lvert x_{i,t+1} - x_{it}\rvert$. The restrictions \eqref{msmk_c4} and \eqref{msmk_c5} enforce that $x_{it}$ and $y_{it}$ are binary variables. 

Solution approaches for traditional multi-dimensional knapsack problems include exact algorithms and heuristic or metaheuristic algorithms, which we refer to the review of \citet{varnamkhasti2012overview} for further details. \citet{bertsimas2002approximate} present an algorithm named adaptive fixing heuristic to solve the general multi-dimensional knapsack problem. Since their heuristic approach achieves good quality solutions \citep{wilbaut2008survey}, we adopt the heuristic presented by \citet{bertsimas2002approximate} to generate a benchmark solution for MSMK. Also, \citet{hill2012problem} present a heuristic that utilizes Lagrangian relaxation to estimate which variables should be fixed.

\section{Methodology}
\label{Methodology}

In this section, we first discuss the machine translation model used to learn the optimal solutions to optimization problems. Here, the encoder-decoder model with the attention adapted from \citet{luong2015effective} is presented with the modifications made. Then, the developed PredOpt framework is introduced. We then present an algorithm for generalization with item-wise expansion.

\subsection{Neural Machine Translation and Adaptation}	
\label{Neural Machine Translation}

Our methodological goal is to create a mapping from input defined by the problem parameters and optimal solutions to an output of predicted optimal solutions to the problem of interest through utilizing neural machine translation frameworks. A machine translation system is used to translate the input sequence $x_{1},x_{2}, \ldots,x_{m}$ from the source language to the output sequence $y_{1},y_{2}, \ldots,y_{n}$ to the target language, where $m$ and $n$ give the length of the input and output sequences, respectively. For optimization problems that we are tackling, the length of the input and output sequences is the same, that is, $m=n$. For MCLSP, the input sequence consists of $d_{it}$, $p_{it}$, $f_{it}$, $h_{it}$, and $c_{t}$, and the output sequence is $y_{it}$ and a binary vector of dimensions $(I+1)T$ that defines tight constraints. For MSMK, the input sequence consists of $p_{it}$, $b_{it}$, $w_{ijt}$, $c_{jt}$, and the output sequence is $x_{it}$ and a $JT$-dimensional binary vector labeling tight constraints.

The main idea of a neural machine translation system is to utilize neural networks to fit a parametrized model to maximize the probability conditioned on the input sequence and the previous output elements: $P (y \mid x) = \prod_{t=1}^{n} P (y_{t} \mid y_{i \mid i < t},x)$. We refer to \citet{stahlberg2020neural} for a detailed review of neural machine translation approaches. Although our neural machine translation architecture is similar to that of \citet{luong2015effective}, we modify their attention mechanism to better accommodate the requirements of the combinatorial optimization problem. At each prediction time step, the hidden states of the top forward and backward LSTM layers, which contain processed input information, are concatenated as presented in \citet{bahdanau2014neural} and used to calculate the attention scores as presented in \citet{luong2015effective}. This is different from that of \citet{luong2015effective}, where they use the hidden states of the top hidden layer to calculate the attention scores.

The encoder is a recurrent neural network that reads the input sequence one at a time. In the context of neural machine translation, at each time step, the encoder reads a word from the input sequence $x_{1},x_{2}, \ldots,x_{m}$. For the encoder, we used bidirectional LSTM networks, which ideally capture the representation of the input sequence by processing with the forward and backward layers. The hidden states of both forward and backward RNNs are used to calculate the attention score and the context to predict the next work in the decoder.

The decoder is also a recurrent neural network that produces the output sequence. At each time step, the decoder input is the prediction from the previous time step. The attention mechanism enables the neural machine translation model to use the information from the encoder hidden states when decoding by determining which parts of the encoder hidden states relate more to the current decoder hidden state. It is a mechanism that enables more focus to be placed on the selected parts of the input sequence. This is done by calculating the so-called context, which is a weighted average of the hidden states of the encoder. The context is used together with the decoder hidden state to output the prediction for each time step. 

The encoder-decoder network with attention is appropriate for making predictions for sequentially dependent optimization problems for various reasons. First and foremost, it can capture the time-wise dependent relationship between the decisions made throughout the problem's horizon. Although the size of the decisions made is constant for each period of the optimization problem, treating these sequential decisions as independent for each period using a classical machine learning model may lead to poorer performance compared to recurrent models \citep{yilmaz2023learning}. Using a sequence-to-sequence mapping model ensures that the knowledge is transferred through the problem's planning horizon. Secondly, recurrence-based attention models have performed extremely well in various tasks, including neural machine translation \citep{bahdanau2014neural}, image captioning \citep{mnih2014recurrent}, and speech recognition \citep{chorowski2015attention}. In \citet{yilmaz2023learning}, the authors use classical LSTMs to predict the optimal solutions. However, an increase in performance in the predictive power of the model can lead to less infeasibility, a smaller optimality gap, and a faster solution time improvement compared to their results. Therefore, by using a more advanced model, solutions to more difficult and complex problems can be learned. Additionally, with the use of the decoder, the decisions made in the previous period are directly considered when predicting the solution for the next period, which is not the case in regular LSTMs. Furthermore, the attention structure enables one to focus on not just the current prediction period's hidden states, but it rather enables one to selectively focus on all periods' hidden states, which is a desirable feature in sequential prediction tasks.

The building blocks of the encoder-decoder model are the LSTMs developed by \citet{hochreiter1997long}. However, the sequence-to-sequence model presented in \citet{luong2015effective}, as described below, better suits our problem compared to the LSTM. First, we consider the monotonic alignment (local) for the attention calculations rather than the global attention scores. In global attention, the context is calculated by taking a weighted average of the hidden states of all periods in the whole input sequence. In the local attention mechanism, a few previous and later periods are considered for the attention score calculation for the period a prediction is made rather than the whole input sequence. In other words, if a prediction is made for period $t$, the attention score; therefore, the context is calculated using the hidden states of the encoder from period $t-D$ to $t+D$, where $D$ represents the size of the window for the local-$D$ attention. 

There are two advantages of this choice. First, the combinatorial optimization problems we are trying to predict can be much longer than the problems used in training. Our predictions benefit from focusing on a time window for production and inventory decisions.  In such cases, the initial periods have little to no effect on the final periods' decisions, considering the length and characteristics of the problems to reset the production amount with each production decision. For example, in MCLSP, once a prediction decision is made, the amount of production is usually such that the demand is fully covered for a few subsequent periods. More often than not, the inventory at the end of the last period is covered with a previous production and reduced to zero or close to zero to avoid inventory holding costs. Inventory reaches zero, and production must be repeated intermittently at some periods throughout the planning horizon rather than only in the last period. In this setting, the period that a prediction is being made has a more intertwined relationship with a few predecessor and successor periods rather than far away periods or the whole sequence. The attention structure enables the model to discover such relationships without explicitly programming them. Second, local$D$ attention is less computationally expensive compared to global attention, and training and prediction times can be reduced by using local$D$ attention compared to full-sequence attention. Further details on encoder-decoder are presented in \ref{Details of the Encoder-Decoder Network}.

\vspace{-0.3cm}
\subsection{The PredOpt Framework}

\noindent \textbf{Training and Validation.} Our prediction approach to learning optimal solutions begins with data generation. We randomly generate all the parameters for the optimization problem of interest. Then, the generated problems are solved using CPLEX. The resulting data set is divided into two, following the standard learning schema \citep{alpaydin2020introduction}: training and validation sets. Both training and validation data sets include the parameters of an instance and its optimal solution for a set of instances. Using the training set, the parameters of the neural machine translation model are optimized to create an output mapping of the optimal solutions for the problem of interest. The models with different hyperparameters are then evaluated using the validation set. A third independent test set of instances with a higher number of periods is generated to measure the effectiveness of the proposed PredOpt framework. 

\noindent \textbf{Testing.} The PredOpt framework presents a strategy to quickly eliminate infeasible predictions. In \citet{yilmaz2023learning}, the predictions of the decision variables are fixed in the solution, and the results show that predictions can cause solutions that are infeasible at a significant rate. \citet{yilmaz2023learning} state that infeasibility depends on the prediction level and can be adjusted empirically. This can be a rather computationally challenging process and does not guarantee the feasibility of the predictions. The main idea in the PredOpt framework is to construct a feasibility-check loop to determine the highest level of predictions that will not lead to an infeasible solution. In the PredOpt framework, the prediction level is not constant for each instance as in \citet{yilmaz2023learning}, but it is determined with a feasibility check loop and a relaxation of the original problem, as described in the next two subsections.

\subsubsection{Predicting Tight Constraints and Forming the Relaxation Problem}
A relaxation of the original problem is formed by identifying the tight and close to tight constraints to reduce the feasibility check time of the PredOpt framework. For brevity, we will refer to those inequalities as tight even if the inequality is not held strictly. The constraints identified as tight are included in the relaxation of the formulation, and those identified as not tight are removed from the relaxation. 

During training, the model is not only used to learn the values of the optimal binary decision variables, but is also used to learn the tight constraints. Thus, before training starts, using the optimal solutions from the training data sets, we determine tight constraints and label them so that the trained model can learn which constraints are tight in the test set. For example, given that $x_{it}^*$ and $y_{it}^*$ are the optimal solutions to an MCLSP \eqref{mclsp_formulation} training instance, constraints \eqref{mclsp_c2} are identified as tight with a predetermined tightness coefficient $\eta \in [0,1]$ if $\sum_{i=1}^{I} x_{it}^* \geq \eta*c_{t}$ and are identified as non-tight otherwise. The constraints \eqref{mclsp_c3} are identified to be tight if $ x_{it}^* \geq \eta*y_{it}^*c_{t}$ and are identified to be non-tight otherwise. For the MSMK presented in formulation \eqref{msmk_formulation}, constraints \eqref{msmk_c1} are tight if $\sum_{i=1}^{I} w_{ijt}x_{it}^* \geq \eta*c_{jt}$ and non-tight otherwise. Constraints \eqref{msmk_c2}, \eqref{msmk_c3}, and \eqref{msmk_c5} are discarded from relaxation since they do not affect the feasibility of the knapsack constraints presented. Relaxation involves binary decision variables for both MCLSP and MSMK. 

As explained above, the full training data set includes both the optimal solutions to the problem and the tight constraints identified from those optimal solutions. In MCLSP, in each period $t$, the network predicts a total of $2\cdot I+1$ variables in each time period, where $I$ of them are used to predict variables $y_{it}$ and the number of variables $I$ is used to predict the tightness of $I$ constraints \eqref{mclsp_c3}. Then the additional variable represents the tightness prediction for the single constraint \eqref{mclsp_c2} at time $t$. Similarly, in MSMK, for each period $t$, a total of $I+J$ predictions are made where the first $I$ represents the decision variables $x_{it}$ and the remaining $J$ predicts whether the constraints \eqref{msmk_c1} are tight or not. Although the sizes of the output dictionaries are $2\cdot I+1$ and $I+J$ for MCLSP and MSMK, respectively, in neural machine translation, at each period a single element is predicted in the output dictionary. In PredOpt, unlike neural machine translation, the problem is a multi-label binary classification problem. Therefore, at each period, a subset of the output dictionary is selected, possibly with multiple elements.

\subsubsection{Feasibility Check and Infeasible Elimination Loop}
\label{Feasibility Check and Infeasible Elimination Loop}

\begin{figure}[h!]

   \centering
   \includegraphics[width=\linewidth]{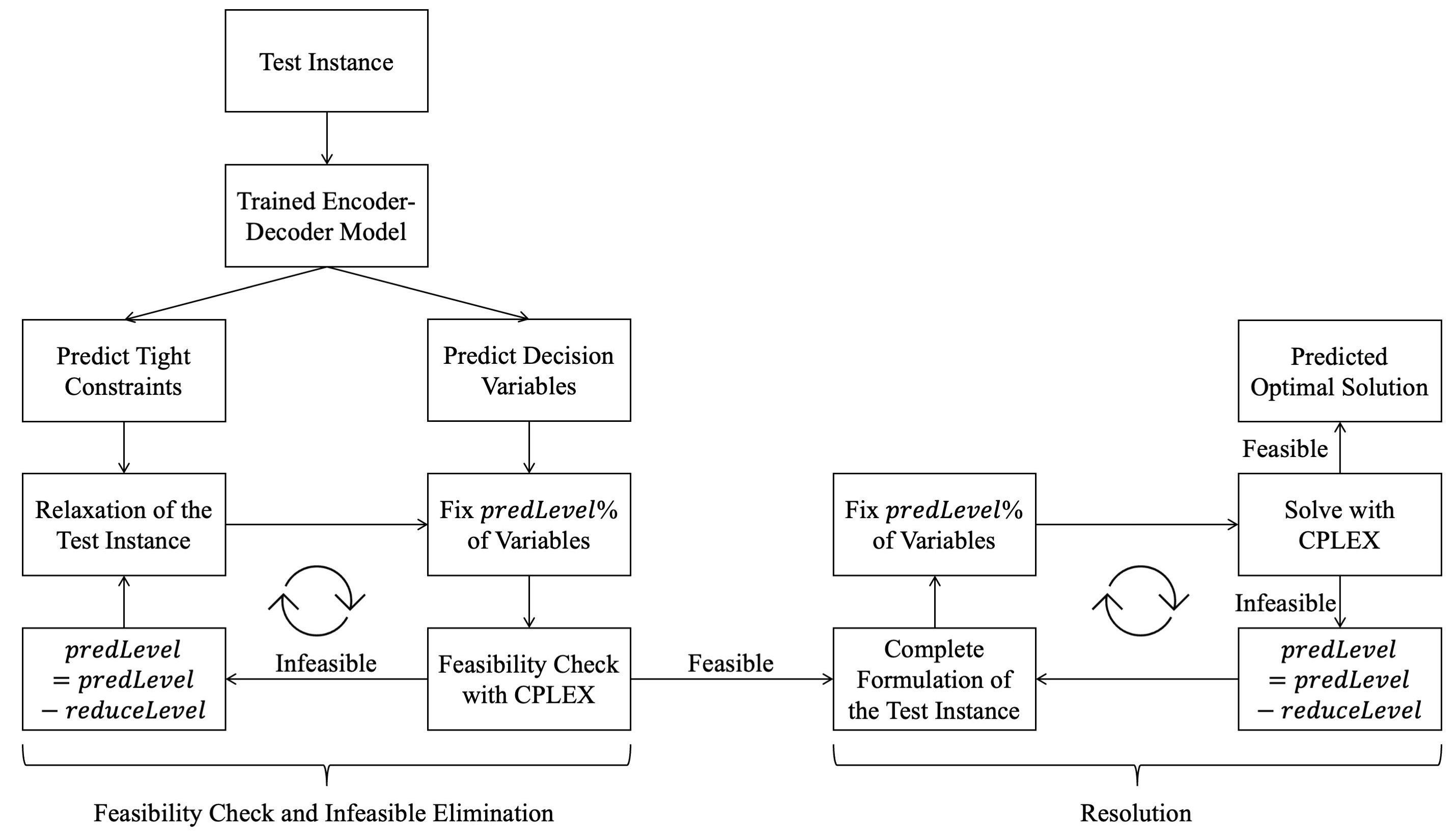}
   \caption{PredOpt Framework}\label{methodologyfigure}
   
\end{figure}

As discussed previously, an attentional encoder-decoder model is used to learn both optimal solutions and tight constraints of the problem of interest. Then, during the testing phase, a strategy is applied to determine the appropriate level of prediction for each instance. Figure \ref{methodologyfigure} presents the PredOpt framework in detail. On the left side of the figure (Feasibility Check and Infeasible Elimination), for each instance in the test set, a set of predictions is generated using the trained network. Using the set of predicted tight constraints, a relaxation of the original model is generated for a fast feasibility check. Then $predLevel\%$ of the predicted variables are fixed in the relaxed formulation. This is done by calculating $maximum(\hat{y},1-\hat{y})$, ordering decreasingly, and taking the top $predLevel\%$ to fix in the model where $\hat{y}$ represents predictions. In this process, the variables are ordered according to their proximity to 0 or 1, and the first $predLevel\%$ of them are fixed in the relaxed problem. If a feasible solution is found by solving the relaxed instance with fixed variables using CPLEX, the loop is exited, and the determined level of prediction is used in the original formulation to determine the values of other variables and the optimal objective function value. If a feasible solution is not found, $predLevel\%$ is reduced by $reduceLevel\%$, which cannot be less than zero, and the feasibility check is repeated with the new $predLevel\%$ until a feasible solution is found.

Once a feasible solution is found, a new loop begins to solve the complete formulation of the test instance, as demonstrated on the right side of Figure \ref{methodologyfigure} (resolution). The determined level of prediction is fixed in the original formulation, which is attempted to be solved with CPLEX. If a feasible solution is found, the loop is exited, and the optimal solution is used to report the quality of the proposed framework. The relaxation of the problem from the first loop of Figure \ref{methodologyfigure} does not necessarily guarantee a prediction level that gives a feasible solution, since it is only a relaxation. Thus, if a feasible solution is not found by solving the original problem with fixed values, similar to the first loop on the left side, $predLevel\%$ is reduced by $reduceLevel\%$, and the loop continues until a feasible solution is found. Computational results show that the number of iterations in the second loop is very few compared to the number of iterations made in the first loop, highlighting the high quality of the relaxation problems in the first loop.

\subsection{Generalization with Item-wise Expansion}
\label{Generalization with Item-wise Expansion}

In this section, we discuss the generalization of trained models to predict longer and higher-dimensional problems. The encoder-decoder can be inherently used to predict for the longer instances than they are trained with, since the size of the predictions is not limited by the length of training instances. The attention structure helps in the collection of encoder information and passes it to the decoder without a limitation on the input length. Therefore, the encoder-decoder with attention is able to keep up with the increasing number of periods, and thus the time-wise expansion of the prediction framework is straightforward. The key question we address here is whether models trained with a few items can be used to predict problems with a large number of items. For example, a significant time reduction can be achieved if an 8-item model can successfully predict a 32-item problem. Therefore, we can solve instances with a small number of items to predict instances with a larger number of items. This would allow us to perform training only once to solve instances generated from the same distribution without retraining if the number of items considered changes. In addition, it can reduce the time to generate training instances and the whole training process.

Here, our strategy includes making multiple forward passes using the trained model with a subset of the items rather than predicting all decision variables at once. In Algorithm \ref{item-wisealg}, we present our strategy for generating predictions for instances with a larger set of items. In Step 1, we initialize the prediction counter for each item $i$, $\gamma_i$, as zero. The algorithm continues until each item $i$ has been predicted at least $\delta =10$ times. At each iteration, we select a subset $S$ of items such that the size of the subset $S$ is equal to $I^M$, representing the number of items with which the model is trained. For example, if a prediction is made using an 8-item model, then only $\lvert S \rvert= 8$ items are selected in Step 3. Then, in Step 4, we make a forward pass with the model $M$ using the input data $\alpha_{S}$ of the selected subset $S$ of the elements to generate predictions $\hat{\theta}_{S}$. Here, the right-hand sides of the constraints are scaled down with the proportion of selected items in order to mimic the original problem. For MCLSP, we modify the right-hand side of constraint \eqref{mclsp_c2} as $c_t=c_t \times { \sum_{i \in S} d_{it} \over \sum_{i=1}^{I} d_{it} },$ $\forall t=1,\ldots,T$. For MSMK, we modify the right-hand side of the constraint \eqref{msmk_c1} as $c_{jt}=c_{jt} \times { \sum_{i \in S} w_{ijt} \over \sum_{i=1}^{I} w_{ijt} },$ $\forall t=1,\ldots,T,\enspace \forall j=1,\ldots,J $. The predictions are saved in Step 5 by summing current predictions $\hat{\theta_i}$ and previous predictions $\hat{\beta_i}$ for each item $i$ in the subset $S$. Then, in Step 6, the count of prediction $\gamma_i$ is increased for each item $i$ in the subset $S$. Finally, the final prediction $\hat{\beta_i}$ for each item $i$ in the entire test set is calculated by dividing the sum of predictions $\hat{\beta_i}$ by their respective counts $\gamma_i$.

\renewcommand{\baselinestretch}{1}
\begin{algorithm}[!htb]

\caption{Item-wise Generalization Prediction Algorithm}\label{item-wisealg}
\hspace*{\algorithmicindent} \textbf{Input:} Trained model $M$, number of items during model training $I^M$, input data of the test set $\alpha$, set of items in the test set $i \in \left\{1,\ldots,I\right\}$, and threshold prediction count $\delta$ \\
\hspace*{\algorithmicindent} \textbf{Output:}  Predicted value for item  $\forall  i \in$ $\left\{1,\ldots,I\right\}$, $\hat{\beta_i}$, for the test set \\
\hspace*{\algorithmicindent} \textbf{Item-wise Generalization} 
\begin{algorithmic}[1]

\STATE Initialize prediction count of each item to be zero: $\gamma_i=0,$ $\forall  i \in$ $\left\{1,\ldots,I\right\}$
\WHILE{$\exists \ \gamma_i$ $\leq$ $\delta $ $i \in$ $\left\{1,\ldots,I\right\}$ }

        \STATE Sample a subset of items, $S$, with the number of items in the subset equal to the number of items in the trained model $I^M$: $S \subset \left\{1,\ldots,I\right\}$ and $	\lvert S \rvert = I^M$
        \STATE Make a forward pass using the input data of the selected subset $S$ of items: $\hat{\theta}_{S} = M(\alpha_{S})$
        \STATE Save predictions for the subset $S$ of selected items: $\hat{\beta_i}:=\hat{\beta_i} + \hat{\theta_i},$  $\forall  i \in$ $S$
        \STATE Increase prediction counts for subset $S$ of selected items: $\gamma_i=\gamma_i+1,$ $\forall i \in S$
\ENDWHILE
\STATE Calculate the final prediction value for each item $i \in$ $\left\{1,\ldots,I\right\}$ by dividing the sum of saved predictions with respective counts: $\hat{\beta_i}:=\hat{\beta_i} \div \gamma_i$.

\end{algorithmic}
\end{algorithm}
\renewcommand{\baselinestretch}{}

\vspace{-0.5cm}

\section{Implementation and Experimentation}	
\label{Implementation}

In this section, we present the details of instance generation, encoder-decoder parameters, and metrics used to measure the performance of the PredOpt framework. The MCLSP and MSMK instances are solved using Python 3.8.5 with DOcplex API and CPLEX 20.1.0. The generation of training and test instances, model training, and testing with PredOpt is completed on a high-performance computing cluster running Linux 3.10.0 with Intel Xeon Gold 6226R 2.90 GHz, 96 GB of memory, and NVIDIA Tesla T4 GPU with Python 3.8.5. The encoder-decoder model is trained with PyTorch 1.7.1 on GPU. The instance generation schemas for MCLSP and MSMK are presented in \ref{Instance Generation}. Model training details are presented in \ref{Model Training Details}.

\subsection{Evaluation Methodology}

In this section, we present the metrics used in evaluating the performance of the PredOpt framework. The metrics are intended to evaluate the success of the PredOpt framework with respect to the optimality gap and the reduction in the solution time for the test set. We follow mainly the success metrics defined in \citet{yilmaz2023learning}:

\begin{itemize}

\item \textbf{timeCPX}: Average solution time of an optimization problem in CPU seconds with CPLEX in the default setting without using any predictions. 
\item \textbf{timePredOpt}: Average solution time of an optimization problem in CPU seconds with CPLEX at the default setting using predictions determined by the PredOpt framework, including prediction generation and infeasibility elimination time.
\item \textbf{timeLS}: Average solution time of an MCLSP in CPU seconds with CPLEX using valid ($\ell$,S) inequalities of \citet{barany1984strong}.
\item \textbf{timeRF}: Average solution time in CPU seconds with the relax-and-fix heuristic of \citet{absi2019worst} for MCLSP.
\item \textbf{timeAC}: Average solution time in CPU seconds with the ant colony heuristic of \citet{almeder2010hybrid} for MCLSP.
\item \textbf{timeLR}: Average solution time in CPU seconds with the Lagrangian Relaxation heuristic of \citet{absi2013heuristics} for MCLSP.
\item \textbf{timeAF}: Average solution time in CPU seconds with the adaptive fixing heuristic of \citet{bertsimas2002approximate} for MSMK.
\item \textbf{timePR}: Average solution time in CPU seconds with the problem-reduction heuristic of \citet{hill2012problem} for MSMK.
\item \textbf{accuracy(\%)}: Percentage of binary variables correctly predicted by the PredOpt framework compared to the optimal solution determined by CPLEX.

\end{itemize}

Also, the following metrics are defined to assess the quality of the PredOpt framework:

\begin{definition}
\label{optgapdef}

Let $x^*$ be the optimal solution of the problem of interest and ${Z}(x^*)$ be the corresponding optimal objective function value. Let $\hat{x}^*$ be the solution determined by the PredOpt framework (or solution determined by a heuristic for calculating $optGapRF (\%)$, $optGapAC (\%)$, $optGapLR (\%)$, $optGapAF (\%)$, and $optGapPR (\%)$) and ${Z}(\hat{x}^*)$ be the corresponding optimal objective function value. The optimality gap is defined as:

\begin{equation} 
\label{equationoptgap}
\mathbf{optGapPredOpt (\%)}  =   \frac{\left|{Z}(\hat{x}^*)-{Z}(x^*)\right|}{{Z}(x^*) } \times 100.
\end{equation} 
\end{definition}

\begin{definition}
\label{improvement}
The solution time improvement factor achieved by the PredOpt framework (timeImpLS for CPLEX with ($\ell$,S) inequalities and timeImpRF, timeImpAC, timeImpLR, timeImpAF, and timeImpPR for heuristic time) with respect to the default CPLEX is given by:

\begin{equation}
\label{equationimprovement}
\mathbf{timeImpPredOpt} = \frac{timeCPX}{timePredOpt}.
\end{equation}
\end{definition}

\begin{definition}
\label{p-value}
The \textbf{p-value} is calculated based on a one-sided Wilcoxon signed-rank test \citep{wilcoxon1945individual}. It is a statistical test that measures if the differences between the two distributions are symmetric around 0. It is a non-parametric version of a paired T-test. The null and alternative hypotheses are:
\begin{subequations}
\begin{eqnarray}
H_{0}: & median(timeRF \enspace(or\enspace timeAF)-timePredOpt) < 0\\
H_{1}: & median(timeRF \enspace(or\enspace timeAF)-timePredOpt) > 0
\end{eqnarray}
\end{subequations}
If the p-value is less than 0.01, the null hypothesis is rejected, implying that $PredOpt$ performs statistically better than the corresponding heuristic in terms of solution time.
\end{definition}

\section{Computational Results}
\label{predoptResults}

In this section, we present the computation results of the PredOpt framework for solving a variety of MCLSP and MSMK instances and compare the results to the direct solution with CPLEX, various heuristics, and an exact approach. For both MCLSP and MSMK, we created 25 different test sets, each consisting of 20 instances, resulting in a total of 1,000 instances. Our MCLSP test instances have up to $T=200$ periods and $I=300$ items. The MSMK test instances contain up to $T=40$ periods and $I=200$ items. The initial prediction level is set at 80\% for MCLSP and 60\% for MSMK, which are effective in estimating the prediction level that eliminates infeasibility. Within the PredOpt framework, the time to generate predictions and find a prediction level that leads to a feasible solution to the relaxed problem is below one second, which is very fast compared to solving the full-sized problem with the determined prediction level. All computational results that refer to solution times are presented in CPU seconds.

We compare the performance of the PredOpt algorithm with the state-of-the-art heuristics. We opted to use heuristics similar in solution approach to our PredOpt framework, which includes relaxations, iterative variable fixing, and mathematical solvers. We employ the relax-and-fix heuristic of \citet{absi2019worst} for MCLSP. In essence, the heuristic iteratively fixes some binary variables and relaxes the remaining ones, similar to how PredOpt iteratively determines which variables should be fixed. In the heuristic of \citet{absi2019worst}, at each iteration, the number of fixed periods is taken as $T \over 20$, and the number of periods with unfixed binary decision variables is set to $T \over 10$. Furthermore, we utilize the metaheuristic approach of \citet{almeder2010hybrid}. It is an ant colony optimization approach that is integrated with CPLEX. Their approach is similar to that of PredOpt in that it combines an existing approach with CPLEX to improve the solution. In addition, we employ the Lagrangian relaxation-based heuristic of \citet{absi2013heuristics}. Their methodology involves relaxing the capacity constraints to solve single-item problems and adapting a dynamic programming algorithm. The solution is then improved with a probing strategy, a refinement procedure, and a neighborhood search. For MSMK, we implement the adaptive fixing heuristic of \citet{bertsimas2002approximate} for comparison with our framework. Similar to PredOpt, it aims to find which binary variables should be fixed based on the relaxation of the problem. Lastly, we make use of the problem-reduction heuristic of \citet{hill2012problem} based on Lagrangian relaxation. This time-competitive heuristic is similar to PredOpt, with the main idea of determining which binary variables to fix and their values.

\vspace{-0.3cm}

\subsection{Quality of Predictions for MCLSP}
\label{Quality of Predictions for MCLSP}

Table \ref{resultmclsp2} presents the results for a set of MCLSP instances. Here, the model is trained only with $T = 40$-period instances, but it is used to test instances from $T = 40$ up to $T = 200$. This shows that the trained models can easily generalize in time dimension with the presented local attention structure. The first set of test instances with $T = 40$ is solved with a mean solution time of 2.3 seconds with CPLEX. In all instances in this test set, the time to generate predictions and the infeasibility elimination loop is fairly short compared to the optimization (resolution) loop of the PredOpt framework shown in Figure \ref{methodologyfigure}. For example, for the first two sets of instances with $T = 40,60$ in Table \ref{resultmclsp2}, the optimal level of predictions that do not cause infeasibility is calculated within 0.01 seconds using the relaxation of the problem. Then, with the prediction level 80\%, this set of instances is solved in a 9-fold faster solution time than CPLEX with only a 0.01\% optimality gap. The PredOpt reaches the same average objective function value over 20 test instances faster compared to the solution strengthened with the ($\ell$,S) inequalities, the relax-and-fix heuristic of \citet{absi2019worst}, ant colony optimization heuristic of \citet{almeder2010hybrid}, or Lagrangian relaxation heuristic of \citet{absi2013heuristics}. We do not report an optimality gap with the ($\ell$,S) inequalities since they solve the test instances to optimality without a solution time limit, but the optimality gap of the heuristic is much higher than the optimality gap of PredOpt.  PredOpt is statistically faster than the relax-and-fix heuristic, since the p-value is smaller than 0.001. Furthermore, both the PredOpt and the relax-and-fix heuristic outperform the ant colony metaheuristic in both solution time and quality. In addition, the PredOpt outperforms the Lagrangian relaxation heuristic in terms of the optimality gap for all test cases.

For all periods, the time of the PredOpt framework is much less than the CPLEX solution time. Time improvements increase as the number of periods increases and problems get harder. For example, in the last data set with $T = 200$, the solution time is reduced from more than 10 minutes to slightly above 3 seconds with an optimality gap of 0.02\%. Also, the accuracy of the models does not deteriorate as the number of periods multiplies, confirming that the encoder-decoder with the local attention structure is able to capture the problem characteristics. The optimality gap is kept below 0.03\% for all cases in the test sets used in Table \ref{resultmclsp2}. 

As a summary of Table \ref{resultmclsp2}, our model trained with shorter periods can predict 5-fold longer problems with the local attention structure. For all test instances, the PredOpt provides a faster solution than the CPLEX, ($\ell$,S) inequalities and the heuristic with an optimality gap better than that of both heuristics. The optimality gap remains constant on all test instances with an average of 0.02\%, and PredOpt captures the characteristics of the problem even if the size of the test instances grows significantly.

\begin{table}[h!]
\caption{Average results of experiments for MCLSP with 8 items}
\label{resultmclsp2}
\centering
\begin{tabularx}{\textwidth}{rCCCCCC}

$T$	&	40	& 	60	&  	80	&   	100	&    	150	&     	200	\\ \hline
timeCPX	&	2.3	&	4.2	&	4.9	&	30.9	&	62.2	&	659.2	\\
timePredOpt	&	0.3	&	0.4	&	0.6	&	1.2	&	2.1	&	3.4	\\
timeLS	&	3.0	&	5.6	&	7.5	&	12.5	&	35.9	&	66.6	\\
timeRF	&	10.6	&	10.5	&	12.0	&	13.1	&	19.9	&	23.7	\\
timeAC	&	12.5	&	17.2	&	22.4	&	27.4	&	38.8	&	48.3	\\
timeLR	&	0.4	&	0.6	&	0.6	&	0.8	&	1.5	&	2.0	\\
timeImpPredOpt	&	9	&	12	&	8	&	22	&	28	&	183	\\
timeImpLS	&	1	&	1	&	1	&	1	&	1	&	6	\\
timeImpRF	&	0	&	0	&	0	&	2	&	3	&	32	\\
timeImpAC	&	0	&	0	&	0	&	1	&	2	&	14	\\
timeImpLR	&	6	&	7	&	8	&	41	&	45	&	335	\\
p-value	&	$ \leq 0.001$	&	$ \leq 0.001$	&	$ \leq 0.001$	&	$ \leq 0.001$	&	$ \leq 0.001$	&	$ \leq 0.001$	\\
accuracy(\%)	&	99.6	&	99.7	&	99.7	&	99.6	&	99.7	&	99.7	\\
optGapPredOpt(\%)	&	0.01	&	0.01	&	0.02	&	0.03	&	0.02	&	0.02	\\
optGapRF(\%)	&	1.08	&	1.11	&	1.03	&	1.08	&	0.41	&	1.05	\\
optGapAC(\%)	&	2.39	&	2.45	&	2.51	&	2.51	&	2.88	&	2.77	\\
optGapLR(\%)	&	1.76	&	1.93	&	1.63	&	1.91	&	2.14	&	1.86	\\ \hline

 \end{tabularx}
\end{table}

Table \ref{resultmclsp3} presents a more difficult set of test instances with 12 items. The model is only trained using $T=30$ and $I=12$ instances, but the results are presented to solve models with periods ranging from $T=30$ to $T=150$. Similar to Table \ref{resultmclsp2}, time improvements improve significantly as the number of periods in the test problem multiplies. For example, the data set with $T=125$ is solved on average for more than 1 hour using CPLEX at the default setting. PredOpt reduces the solution time to less than 7 seconds with an optimality gap below 0.1\% without infeasibility in the test set. For the last data set with $T=150$, the solution time is reduced from almost 7 hours to just under 5 seconds with an optimality gap of 0.11\%. This translates into a reduction in solution time of a factor of 7,236. For all test sets, accuracy is maintained as the time period increases. Also, the p-value for the Wilcoxon signed-rank test is fairly small with a lower optimality gap compared to the relax-and-fix heuristic for all cases, showing that the PredOpt reaches a better objective function value in a shorter time than the heuristic. While ($\ell$,S) inequalities are effective in reducing the solution time for harder instances, they are still significantly slower than the PredOpt at the cost of a very small optimality gap. 

To sum up the results in Table \ref{resultmclsp3}, the increase in time improvement factors increases significantly for the harder problems. The optimality gap stays quite steady as we increase the number of periods in the test set, which highlights the potential of the PredOpt framework to successfully predict for much longer instances. Also, the PredOpt framework is faster and better than the relax-and-fix and ant colony heuristics and has a much lower optimality gap compared to the Lagrangian relaxation heuristic. We present a detailed comparison of our framework and CPLEX with ($\ell$,S) inequalities in terms of reducing the objective value faster in \ref{PredOpt Reduces Objective Value Faster than CPLEX}.

\begin{table}[h!]
\caption{Average results of experiments for MCLSP with 12 items}
\label{resultmclsp3}
\centering
\begin{tabularx}{\textwidth}{rCCCCCC}

$T$	&	30	& 	50	&  	75	&   	100	&    	125	&     	150	\\ \hline
timeCPX	&	1.1	&	10.3	&	80.9	&	86.8	&	3982.5	&	25085.3	\\
timePredOpt	&	0.3	&	0.6	&	1.5	&	2.0	&	6.3	&	4.6	\\
timeLS	&	2.4	&	5.9	&	12.0	&	18.2	&	2259.4	&	53.1	\\
timeRF	&	18.0	&	14.9	&	17.5	&	18.7	&	22.2	&	27.6	\\
timeAC	&	15.2	&	26.8	&	31.2	&	40.6	&	50.1	&	52.3	\\
timeLR	&	0.4	&	0.5	&	0.8	&	1.1	&	1.5	&	2.2	\\
timeImpPredOpt	&	4	&	14	&	47	&	40	&	696	&	7236	\\
timeImpLS	&	0	&	2	&	4	&	5	&	49	&	453	\\
timeImpRF	&	0	&	1	&	4	&	5	&	172	&	867	\\
timeImpAC	&	0	&	0	&	3	&	2	&	85	&	470	\\
timeImpLR	&	3	&	19	&	101	&	77	&	2371	&	11948	\\
p-value	&	$ \leq 0.001$	&	$ \leq 0.001$	&	$ \leq 0.001$	&	$ \leq 0.001$	&	$ \leq 0.001$	&	$ \leq 0.001$	\\
accuracy(\%)	&	99.5	&	99.3	&	99.2	&	99.0	&	99.0	&	98.9	\\
optGapPredOpt(\%)	&	0.04	&	0.05	&	0.06	&	0.10	&	0.09	&	0.11	\\
optGapRF(\%)	&	0.52	&	0.86	&	0.83	&	1.14	&	0.87	&	0.43	\\
optGapAC(\%)	&	1.91	&	2.20	&	2.12	&	2.07	&	2.16	&	2.22	\\
optGapLR(\%)	&	2.32	&	2.01	&	2.07	&	1.87	&	2.03	&	2.55	\\ \hline

\end{tabularx}
\end{table}

\subsection{Quality of Predictions for MSMK}

\label{Quality of Predictions for MSMK}

This section presents the results for the MSMK, similar to Section \ref{Quality of Predictions for MCLSP}. Table \ref{resultmsmk0} presents the results for the model trained with the 8 items, 30 periods, and 5 resource constraints. The model is used to predict the optimal solution of instances with a wide range of periods from $T=30$ to $T=200$. Unlike Section \ref{Quality of Predictions for MCLSP}, we do not compare with an exact algorithm like ($\ell$,S) inequalities but compare the solution performance with the CPLEX performance and the adaptive fixing heuristic of \citet{bertsimas2002approximate} and the problem-reduction heuristic of \citet{hill2012problem}. In Table \ref{resultmsmk0}, the solution time for 30-period instances is reduced by a factor of 6 with respect to the CPLEX solution time. As instances become harder, the time improvement increases significantly. For the last data set with 200 periods, the solution time is reduced from more than 3.5 hours to 3 seconds with four orders of magnitude reduction of the solution time by PredOpt over CPLEX. The accuracy of the variables selected by the PredOpt framework is somewhat lower than the MCLSP instances presented in Tables \ref{resultmclsp2} and \ref{resultmclsp3}. Therefore, the resulting optimality gaps are at a slightly higher level. We believe that the harder combinatorial nature of the MSMK possesses a larger learning challenge. Additionally, the existence of multiple tighter constraints with pure binary variables per period for MSMK, i.e., Equation \eqref{msmk_c1}, compared to a single binding constraint per period involving continuous variables, i.e., Equation \eqref{mclsp_c2}, for MCLSP, makes MSMK a more challenging problem to predict. However, the optimality gaps are still below 0.98\%, which is a much better solution performance than the heuristic, which gives an optimality gap greater than 2. 7\%. Also, the PredOpt achieves a better objective function statistically faster than the adaptive fixing heuristic, since all p values for the Wilcoxon signed-rank test are smaller than 0.001. In addition, PredOpt outperforms the problem reduction heuristic for all instances in solution time and 3 out of 6 instances in terms of the optimality gap.

As a summary of Table \ref{resultmsmk0}, the PredOpt framework outperforms the heuristic in both the time and the optimality gap aspects. The solution time is reduced by four orders of magnitude with respect to CPLEX and one to two orders of magnitude with respect to the heuristic. This shows that our framework is beneficial for dealing with instances that are difficult to solve.

\begin{table}[h!]
\caption{Average results of experiments for MSMK with 8 items}
\label{resultmsmk0}
\centering
\begin{tabularx}{\textwidth}{rCCCCCC}

$T$	&	30	& 	50	&  	80	&   	100	&    	150	&     	200	\\ \hline
timeCPX	&	0.9	&	4.1	&	67.7	&	772.8	&	4579.8	&	12862.5	\\
timePredOpt	&	0.2	&	0.2	&	0.4	&	2.4	&	0.5	&	2.7	\\
timeAF	&	7.7	&	14.1	&	24.2	&	43.2	&	73.6	&	121.3	\\
timePR	&	1.1	&	2.0	&	7.6	&	47.1	&	233.5	&	403.8	\\
timeImpPredOpt	&	6	&	22	&	207	&	2854	&	12449	&	15218	\\
timeImpAF	&	0	&	0	&	3	&	17	&	62	&	107	\\
timeImpPR	&	1	&	2	&	8	&	37	&	40	&	58	\\
p-value	&	$ \leq 0.001$	&	$ \leq 0.001$	&	$ \leq 0.001$	&	$ \leq 0.001$	&	$ \leq 0.001$	&	$ \leq 0.001$	\\
accuracy(\%)	&	92.9	&	93.6	&	92.9	&	91.6	&	92.3	&	93.8	\\
optGapPredOpt(\%)	&	0.75	&	0.67	&	0.80	&	0.91	&	0.98	&	0.71	\\
optGapAF(\%)	&	2.74	&	2.92	&	3.07	&	3.01	&	2.87	&	2.94	\\
optGapPR(\%)	&	2.30	&	1.23	&	1.14	&	0.89	&	0.50	&	0.54	\\ \hline

\end{tabularx}
\end{table}

In Table \ref{resultmsmk1}, results for the model trained with the 10 items, 30 periods, and 4 resource constraints are presented. These results show similarities with the results presented in Table \ref{resultmsmk0}. The solution times are significantly reduced for all test instances with periods from $T=30$ to $T=100$. In the final data set with $T=100$, the solution time is limited by four hours with CPLEX, resulting in an optimality gap of 0.002\%. The solution time is reduced from 4 hours to under a second with an optimality gap of 0.68\%, resulting in a very significant reduction in solution time. For the same data set, the adaptive fixing heuristic can only achieve a solution with an optimality gap of 2. 4\% in 4 seconds, and the problem reduction heuristic reduces the solution time to 171 seconds with an optimality gap of 1.29\%. To sum up, Table \ref{resultmsmk1}, the p-values for the Wilcoxon signed-rank test are very small for the instances that are hard to solve, ensuring that the PredOpt framework performs better than the adaptive fixing heuristic in both the time and optimality gap aspects. Also, PredOpt achieves a better optimality gap than the problem reduction heuristic in 5 over 6 instances with the shortest solution times.

\begin{table}[!htbp]
\caption{Average results of experiments for MSMK with 10 items}
\label{resultmsmk1}
\centering
\begin{tabularx}{\textwidth}{rCCCCCC}


$T$	&	30	& 	50	&  	70	&   	80	&    	90	&     	100	\\ \hline
timeCPX	&	2.5	&	41.7	&	1051.4	&	4408.8	&	9057.8	&	14409.7	\\
timePredOpt	&	0.2	&	0.4	&	0.5	&	0.5	&	52.5	&	0.7	\\
timeAF	&	8.0	&	14.9	&	22.2	&	30.9	&	34.2	&	40.9	\\
timePR	&	1.3	&	5.2	&	24.9	&	132.8	&	255.3	&	171.4	\\
timeImpPredOpt	&	14	&	170	&	2027	&	8673	&	12580	&	28451	\\
timeImpAF	&	0	&	3	&	49	&	138	&	262	&	355	\\
timeImpPR	&	2	&	17	&	52	&	492	&	222	&	803	\\
p-value	&	$ \leq 0.001$	&	$ \leq 0.001$	&	$ \leq 0.001$	&	$ \leq 0.001$	&	$ \leq 0.001$	&	$ \leq 0.001$	\\
accuracy(\%)	&	93.0	&	92.5	&	92.4	&	93.1	&	91.2	&	93.2	\\
optGapPredOpt(\%)	&	0.80	&	0.81	&	0.67	&	0.72	&	1.29	&	0.68	\\
optGapAF(\%)	&	2.46	&	2.37	&	2.35	&	2.49	&	2.51	&	2.40	\\
optGapPR(\%)	&	2.46	&	2.28	&	1.11	&	1.22	&	0.98	&	1.29	\\ \hline

\end{tabularx}
\end{table}

\vspace{-0.3cm}

\subsection{Generalization: Quality of Predictions with Item-wise Expansion}

In this section, we present results for item-wise generalization by applying Algorithm \ref{item-wisealg} presented in Section \ref{Generalization with Item-wise Expansion}. Table \ref{resultmclspgen2} presents the results of the item-wise generalization for MCLSP. The table contains results with test instances up to $T=100$ and $I=160$. For MCLSP generalization instances, the capacity-to-demand ratio $c$ is increased proportionality with an increasing number of items to prevent infeasibility while generating test instances. The results in Table \ref{resultmclspgen2} are solved by using the trained model with $T=40$ and $I=8$. For example, the model trained with 8 items and 40 periods is used to predict the test set with 32 items and 200 periods in the first column of results. Using Algorithm \ref{item-wisealg} within the PredOpt, the solution time is reduced by a factor of 62 with only a 0.01\% optimality gap. The overall results show similarities with the tables previously presented. The time improvements get better as problems get harder. For the last three datasets, we set capacity-to-demand ratios $c=90,135$ and $180$, respectively, to reduce the effect of increasing capacities and computational complexity. Otherwise, test instances are solved in a shorter time as the number of items increases due to increased capacity. For example, the third and fourth sets have the same number of test items and periods, while the former set has underlying $c=100$, and the latter is sampled using $c=90$. The reduced capacity has significantly increased the solution time from 28 seconds to 2663 seconds.

The last three sets of instances in Table 5 contain significantly more items than the model has trained with. In the fourth data set, the solution time is reduced from more than 55 minutes to 30 seconds with an optimality gap of only 0.04\%. This translates into a reduction in the solution time of 91. Also, all p-values are less than 0.01, ensuring that PredOpt can achieve a statistically faster solution time compared to the relax-and-fix heuristic of \citet{bertsimas2002approximate}. In addition, for all instances, the solution time and optimality gap of PredOpt is smaller than that of the ant colony metaheuristic of \citet{almeder2010hybrid}. Also, while having a faster average solution time, the optimality gaps for the Lagrangian relaxation heuristic of \citet{absi2013heuristics} are much higher compared to PredOpt, making PredOpt a preferable candidate where solution quality is preferred over solution time. These results highlight that the PredOpt framework can be successfully used to predict instances with longer planning horizons and a much larger number of items, which are computationally complex to solve.

\begin{table}[h!]
\caption{Average results of item-wise generalization experiments for the MCLSP model trained with 8 items}
\label{resultmclspgen2}
\centering
\begin{tabularx}{\textwidth}{rCCCCCC}

Test Items	&	32	&	40	&	80	&	80	&	120	&	160	\\
$T$	&	200	& 	150	&  	100	&   	100 	&    	150 	&     	200 	\\
$c$	&	40	& 	50	&  	100	&   	90	&    	135	&     	180	\\ \hline
timeCPX	&	1045.1	&	73.2	&	27.9	&	2663.2	&	1407.8	&	933.5	\\
timePredOpt	&	17.5	&	9.6	&	6.3	&	30.3	&	225.0	&	378.8	\\
timeLS	&	290.3	&	95.9	&	52.4	&	58.7	&	303.4	&	801.2	\\
timeRF	&	36.2	&	34.6	&	38.3	&	50.9	&	435.1	&	1650.5	\\
timeAC	&	181.3	&	165.6	&	211.7	&	214.3	&	426.6	&	769.6	\\
timeLR	&	5.5	&	4.8	&	4.8	&	5.2	&	15.8	&	30.7	\\
timeImpPredOpt	&	62	&	7	&	5	&	91	&	6	&	2	\\
timeImpLS	&	3	&	1	&	1	&	48	&	5	&	1	\\
timeImpRF	&	28	&	2	&	1	&	49	&	3	&	1	\\
timeImpAC	&	6	&	0	&	0	&	12	&	3	&	1	\\
timeImpLR	&	192	&	15	&	6	&	504	&	85	&	30	\\
p-value	&	$ \leq 0.001$	&	$ \leq 0.001$	&	$ \leq 0.001$	&	$ \leq 0.001$	&	$ \leq 0.001$	&	$ \leq 0.001$	\\
accuracy(\%)	&	99.8	&	99.8	&	99.9	&	99.4	&	99.5	&	99.5	\\
optGapPredOpt(\%)	&	0.01	&	0.01	&	0.01	&	0.04	&	0.04	&	0.04	\\
optGapRF(\%)	&	1.18	&	0.45	&	1.20	&	1.20	&	0.45	&	1.20	\\
optGapAC(\%)	&	2.22	&	2.17	&	2.20	&	2.08	&	4.24	&	4.30	\\
optGapLR(\%)	&	0.42	&	0.61	&	0.60	&	1.04	&	0.88	&	0.65	\\ \hline

\end{tabularx}
\end{table}

Table \ref{resultmclspgen3} presents another set of results for MCLSP with the item-wise generalization algorithm. Here, the model trained with $T=30$ and $I=12$ is utilized to solve MCLSP instances. The presented instances contain up to $T=200$ and $I=300$. Similarly to Table \ref{resultmclspgen2}, we set the capacity-to-demand ratios to $c=130,195,260$, and $325$, for the last four sets of instances, respectively, rather than substantially increasing the capacity to obtain difficult instances with significantly more items. The results confirm the computational success of PredOpt, similar to the tables presented previously. For example, in the fourth dataset, PredOpt reduces the solution time from more than one hour to less than one minute with a gap of 0.05\% from the best CPLEX solution. The results show that PredOpt outperforms the relax-and-fix heuristic and the ant colony metaheuristic in terms of both solution time and quality and outperforms the Lagrangian heuristic in terms of solution quality. Also, PredOpt scales much better for larger instances with more items compared to the ant colony approach.

\begin{table}[h!]
\caption{Average results of item-wise generalization experiments for the MCLSP model trained with 12 items}
\label{resultmclspgen3}
\centering
\begin{tabularx}{\textwidth}{rCCCCCCC}

Test Items	&	36	&	48	&	60	&	120	&	180	&	240	&	300	\\
$T$	&	125	& 	100	&  	80	&   	100 	&    	150	&     	200	&      	160	\\
$c$	&	42	& 	56	&  	70	&   	130	&    	195	&     	260	&      	325	\\ \hline
timeCPX	&	4624.2	&	1020.6	&	115.8	&	4313.2	&	2623.3	&	1982.6	&	1726.6	\\
timePredOpt	&	12.4	&	13.0	&	8.9	&	51.5	&	515.1	&	406.4	&	333.5	\\
timeLS	&	68.4	&	46.5	&	31.8	&	98.0	&	634.3	&	1390.4	&	1237.0	\\
timeRF	&	28.1	&	27.6	&	26.6	&	79.3	&	1252.8	&	2984.3	&	2893.7	\\
timeAC	&	133.6	&	145.0	&	133.0	&	307.1	&	659.1	&	1157.4	&	1160.0	\\
timeLR	&	3.2	&	3.1	&	2.9	&	9.7	&	30.0	&	61.8	&	59.9	\\
timeImpPredOpt	&	406	&	73	&	11	&	85	&	5	&	5	&	6	\\
timeImpLS	&	74	&	23	&	3	&	47	&	5	&	1	&	1	\\
timeImpRF	&	163	&	35	&	4	&	54	&	2	&	1	&	1	\\
timeImpAC	&	34	&	7	&	1	&	14	&	4	&	2	&	1	\\
timeImpLR	&	1418	&	331	&	40	&	438	&	86	&	32	&	28	\\
p-value	&	$ \leq 0.001$	&	$ \leq 0.001$	&	$ \leq 0.001$	&	$ \leq 0.001$	&	$ \leq 0.001$	&	$ \leq 0.001$	&	$ \leq 0.001$	\\
accuracy(\%)	&	99.3	&	99.5	&	99.6	&	99.4	&	99.3	&	99.3	&	99.3	\\
optGapPredOpt(\%)	&	0.05	&	0.03	&	0.03	&	0.05	&	0.05	&	0.05	&	0.05	\\
optGapRF(\%)	&	0.94	&	1.18	&	1.20	&	1.20	&	0.45	&	1.20	&	1.20	\\
optGapAC(\%)	&	2.08	&	2.06	&	2.07	&	2.27	&	4.21	&	11.00	&	9.62	\\
optGapLR(\%)	&	0.76	&	0.96	&	1.04	&	1.22	&	1.16	&	0.87	&	0.96	\\ \hline

\end{tabularx}
\end{table}

Table \ref{resultmsmkgen0} presents a set of results for the item-wise expansion results for the MSMK, which are calculated using the trained model used in Table \ref{resultmsmk0}. Here, we reduce the number of periods as we increase the number of items for the first three sets of instances to test the limits of our periodic attention-based learning framework. We consider test instances with parameters up to $T=35$ and $I=160$. Compared to item-wise generalization of MCLSP, the MSMK has a slightly growing optimality gap as the number of items increases and the period decreases. For example, the third data set with 32 items achieves an optimality gap of 1.49\% with the PredOpt framework and the model trained with 8 items. However, this result is obtained in a very small fraction of the CPLEX solution time, that is, 74 minutes of CPLEX compared to 0.3 seconds with PredOpt. Additionally, the last three instances are more challenging due to their large scale, which is reflected in the 2-hour limited solution time. For those instances, PredOpt scales much better than both of the heuristics in solution time and quality. For all experiments, PredOpt is significantly faster than the adaptive heuristic since all p-values are smaller than 0.01. Also, the optimality gaps of PredOpt are lower than those of both heuristics, except for the third data set. The increased optimality gap in the third data set occurs because a lower number of periods in the test set reduces the effectiveness of the attention mechanism and the increasing number of items reduces the success of the item-wise generalization algorithm. On the other hand, PredOpt performs better than CPLEX and the heuristic for all instances in terms of the solution time. These results show that PredOpt is a good alternative to exact solvers and heuristics to use when a fast and accurate solution is needed. Another set of generalization experiment for MSMK is presented in \ref{appendixresultmsmkgen1} outlining a similar result for the success of the PredOpt.

\begin{table}[h!]
\caption{Average results of item-wise generalization experiments for the MSMK model trained with 8 items}
\label{resultmsmkgen0}
\centering
\begin{tabularx}{\textwidth}{rCCCCCC}

Test Items	&	16	&	24	&	32	&	80	&	120	&	160	\\
$T$	&	30	& 	20	&  	15	&   	35	&    	30 	&     	25	\\ \hline
timeCPX	&	466.1	&	827.6	&	4494.3	&	7214.7	&	7211.0	&	7213.3	\\
timePredOpt	&	0.4	&	0.3	&	0.3	&	1.4	&	2.0	&	1.8	\\
timeAF	&	9.5	&	7.5	&	5.8	&	30.0	&	32.5	&	28.4	\\
timePR	&	4.2	&	6.4	&	1.8	&	529.7	&	577.9	&	589.4	\\
timeImpPredOpt	&	1130	&	3149	&	14728	&	5110	&	3636	&	4064	\\
timeImpAF	&	49	&	104	&	780	&	242	&	225	&	257	\\
timeImpPR	&	197	&	288	&	3896	&	33	&	29	&	13	\\
p-value	&	$ \leq 0.001$	&	$ \leq 0.001$	&	$ \leq 0.001$	&	$ \leq 0.001$	&	$ \leq 0.001$	&	$ \leq 0.001$	\\
accuracy(\%)	&	91.2	&	90.7	&	88.7	&	88.1	&	88.2	&	87.5	\\
optGapPredOpt(\%)	&	0.93	&	1.18	&	1.49	&	1.76	&	1.80	&	1.94	\\
optGapAF(\%)	&	1.88	&	1.33	&	1.14	&	4.04	&	3.84	&	3.65	\\
optGapPR(\%)	&	2.70	&	3.72	&	6.07	&	3.76	&	3.83	&	4.84	\\ \hline

\end{tabularx}
\end{table}
\renewcommand{\baselinestretch}{}

\vspace{-0.5cm}

\section{Conclusions and Future Work}
\label{predoptConclusions}		

\vspace{-0.3cm}
In this study, we present a learning-based framework to solve sequential decision-making problems. Our PredOpt framework can help solve optimization problems in an industrial setting where problems with similar structures need to be solved frequently with slightly different parameters. We proposed a learning framework based on a neural machine translation architecture. In the PredOpt framework, we present a strategy to eliminate any infeasible predictions of decision variables in a fraction of a second. To achieve that, a fast feasibility check is performed with a relaxed and smaller problem, which is also generated by using the same trained neural network. Once the best prediction level is determined, the problem is solved in a way that integrates with the commercial solver. Also, we have shown that the models trained on shorter-period problems can be successfully used to predict instances that have multiple times longer periods using a local attention mechanism. The results show that the solution time can be reduced by up to three orders of magnitude with an optimality gap below 0.1\%. In addition, a statistical test confirms that the solution time of the PredOpt framework is faster than that of CPLEX enhanced with valid inequalities specially designed for the problem. Furthermore, we develop and implement an item-wise generalization algorithm and show that the models trained on a small number of items can predict instances with a much larger number of items. Computational results comparing our PredOpt framework with various specially-designed heuristics show that PredOpt outperforms them. The literature includes a large number of heuristics for the lot-sizing and knapsack problems (see, e.g., \cite{buschkuhl2010dynamic} and \cite{freville2004multidimensional}), and each heuristic has pros and cons. The ones that speed up the solution have larger optimality gaps, and the ones that have acceptable optimality gaps run slower. In this paper, we have implemented various types of heuristics and exact approaches to compare them with PredOpt. In most cases, PredOpt outperforms exact approaches in terms of the solution time, while surpassing the quality of the solution provided by all the heuristics used for comparison. These results show the remarkable potential of such ML-Optimization algorithms to solve combinatorial problems over both heuristics and exact solution algorithms.

The length of the planning horizon is often specific to the application considered, and it has been widely studied in the lot-sizing literature \citep{jans2008modeling, buschkuhl2010dynamic, tiacci2012demand, jing2019forecast}. In general, combinatorial problems with medium-horizon problems are shown to be superior to single-horizon problems (see, e.g., \cite{ huang2009value}), and as such, long-period problems can be superior to medium-horizon problems in terms of decision quality. This difference could be especially true for the decisions corresponding to the first few periods, when the problem solution is sensitive to the problem parameters. Lot-sizing problems with long planning horizons are commonly used in practical applications, including large-scale semiconductor manufacturing, energy production, and biofuel production \citep{uzsoy1992review, shrouf2015energy,kantas2015multi}. Such problems include daily or even hourly planning; therefore, a problem with a large planning horizon is required. Even for these problems, the solution might be limited to a few periods due to the computational complexity, although the short-period problem might not capture all the problem details, interrelations, and complexity. Our study aims to close this gap by significantly reducing solution times for long problems to practical levels, thereby enabling the utilization of long-duration problems when they offer a superior solution.

Our PredOpt framework shows a promising direction in the integration of learning-based frameworks with state-of-the-art commercial solvers.  Learners can be used to make easier decisions, and harder decisions can be left to exact approaches to solve. In this spirit, future studies that integrate ML together with traditional OR approaches have the potential to bring out the best of both worlds. Also, we have shown that ML has a potential for generalization from different directions, such as time and number of items. Further studies can focus on learning from solutions of smaller and easier problems and extend this knowledge to solve larger and harder problems. Moreover, new methodologies based on the item-wise generalization algorithm as presented in this paper or reinforcement learning similar to the studies of \cite{yilmaz2023a} and \cite{bushajbuyuktahtakin2023} can be developed and integrated into our framework to solve problems that are initially large. Another future direction could be to further generalize the PredOpt framework to predict instances with different underlying distributions. Additionally, the PredOpt framework can be used to solve various types of sequential problems, including traveling salespeople and vehicle routing.

\noindent \textbf{Declarations of interest} None. \\
\noindent \textbf{Acknowledgment} 
We gratefully acknowledge the support of the National Science Foundation CAREER Award co-funded by the CBET/ENG Environmental Sustainability program and the Division of Mathematical Sciences in MPS/NSF under Grant No. CBET-1554018. We also acknowledge the support of New Jersey Institute of Technology Academic \& Research Computing Systems for High Performance Computing resources. Also, we thank two reviewers and the editor for their helpful feedback, which helped improve our exposition's clarity.

\vspace{-0.3cm}

\bibliographystyle{elsarticle-harv}
\bibliography{PredOptEJOR}

\newpage

\appendix

\section{Details of the Encoder-Decoder Network}
\label{Details of the Encoder-Decoder Network}

\begin{figure}[!htb]

   \centering
   \includegraphics[width=\linewidth]{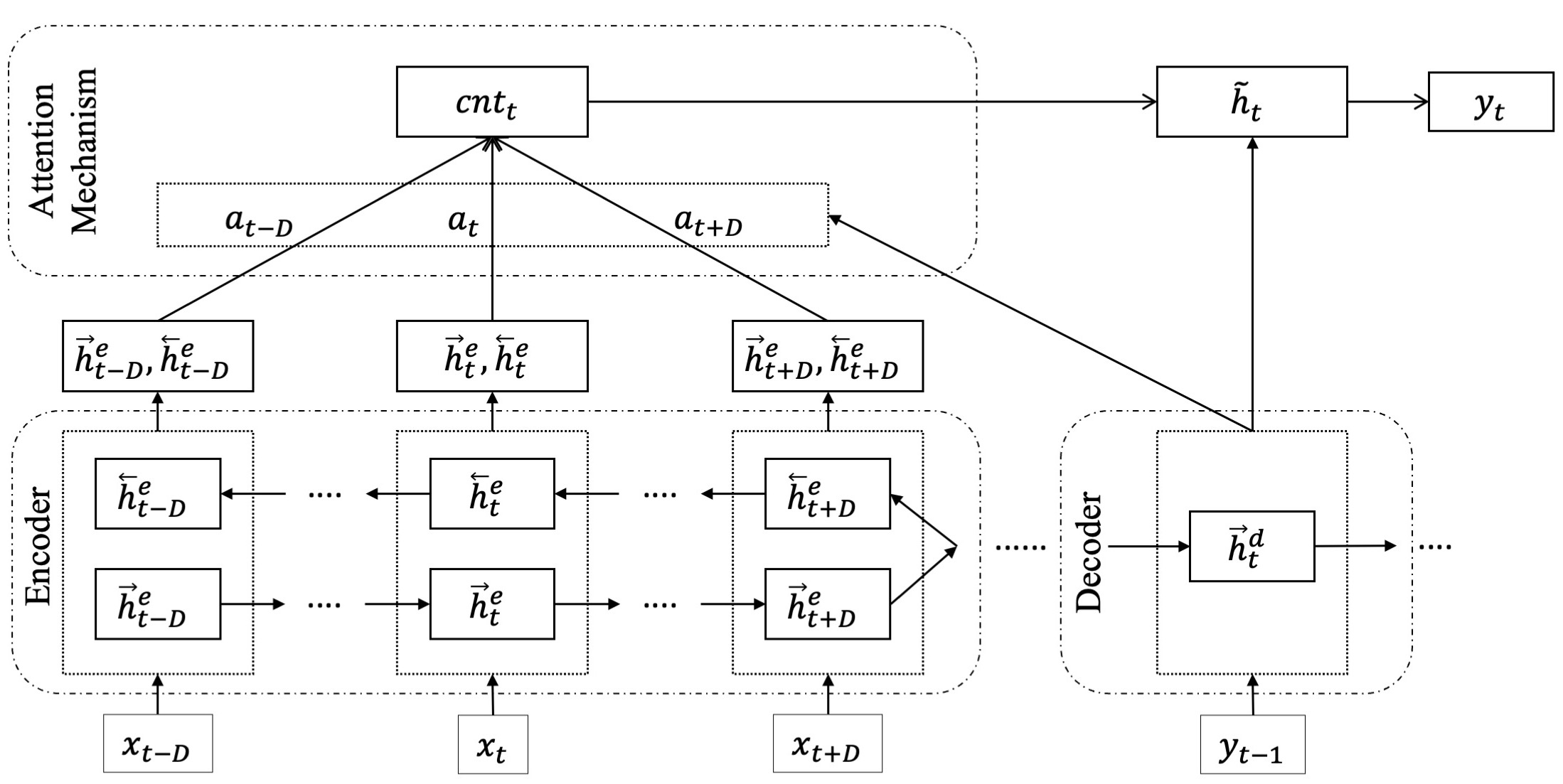}
   \caption{Encoder-decoder with Attention}\label{endoderdecoder}
   
\end{figure}

Figure \ref{endoderdecoder} shows an encoder-decoder model with attention used for a period $t$.  The encoder takes the input sequence $x_{1},x_{2}, \ldots,x_{m}$ but focuses on the sequence $x_{t-D},\ldots,x_{t}, \ldots,x_{t+D}$ in time $t$ with a time window of size $D$. If the boundaries of the input sequence exceed the problem's time horizon, the part outside the windows is ignored. The entire input sequence is processed simultaneously with the forward and backward LSTM layers to generate hidden states, but for the attention calculation, only a portion of this series is considered for the current period $t$. For a period $t$, the hidden state of the encoder generated by the forward layer $LSTM^{e}_{forward}$ is denoted by $\overrightarrow{h}^{e}_{t}$, and the hidden state of the encoder generated by the backward layer $LSTM^{e}_{backward}$ is denoted by $\overleftarrow{h}^{e}_{t}$. Hidden states are the output vectors of LSTM that carry high-level processed information related to the current period. These hidden states are calculated as:

\begin{subequations}
\label{EncoderLSTMlayers}
\begin{eqnarray}
\overrightarrow{h}^{e}_{t-D},\ldots,\overrightarrow{h}^{e}_{t},\ldots,\overrightarrow{h}^{e}_{t+D} = LSTM^{e}_{forward}(x_{t-D},\ldots,x_{t}, \ldots,x_{t+D}) \label{EncoderLSTMlayersforward}\\
\overleftarrow{h}^{e}_{t-D},\ldots,\overleftarrow{h}^{e}_{t},\ldots,\overleftarrow{h}^{e}_{t+D} = LSTM^{e}_{backward}(x_{t-D},\ldots,x_{t}, \ldots,x_{t+D}) \label{EncoderLSTMlayersbackward}
\end{eqnarray}
\end{subequations}
For current prediction period $t$, the decoder hidden state $\overrightarrow{h}^{d}_{t}$ is generated with the output $y_{t-1}$ from previous period $t-1$ using the decoder network $LSTM^{d}$:
\begin{equation}
\overrightarrow{h}^{d}_{t} = LSTM^{d}(y_{t-1}) \label{LSTMlayersforward}
\end{equation}

Then a comparison is made between the current hidden state of the decoder $\overrightarrow{h}^{d}_{t}$ with all encoder hidden states in the attention window $t-D$ to $t+D$. The hidden states of the encoder of the forward and backward LSTM layers of each period $t$ in the attention windows are concatenated as $\overrightarrow{h}^{e}_{t},\overleftarrow{h}^{e}_{t}$ to calculate the attention scores together with the hidden state of the decoder $\overrightarrow{h}^{d}_{t}$ as given below:
\begin{subequations}
\label{attentionscore}
\begin{eqnarray}
a_{i}=\frac{exp(score(\overrightarrow{h}^{d}_{t},[\overrightarrow{h}^{e}_{i},\overleftarrow{h}^{e}_{i}]))}{\sum_{t^{\prime}=t-D}^{t+D} exp(score(\overrightarrow{h}^{d}_{t},[\overrightarrow{h}^{e}_{t^{\prime}},\overleftarrow{h}^{e}_{t^{\prime}}]))} \quad \forall i=t-D,\ldots, t+D. \quad\label{attentionscoreeq}
\end{eqnarray}
\end{subequations}
where scores are calculated as:
\begin{subequations}
\label{attentiongeneral}
\begin{eqnarray}
score(\overrightarrow{h}^{d}_{t},[\overrightarrow{h}^{e}_{i},\overleftarrow{h}^{e}_{i}])={\overrightarrow{h}^{d}_{t}}^\intercal W_{\alpha} [\overrightarrow{h}^{e}_{i},\overleftarrow{h}^{e}_{i}],
\end{eqnarray}
\end{subequations}
where $W_{\alpha}$ is a vector of learned parameters to calculate attention.

After the scores are calculated for each period in the attention window, the context $cn_{t}$, which ideally accumulates all relevant information from the input sequence, is calculated by taking a weighted average: $cn_{t} = \sum_{i=t-D}^{t+D} a_{i} *[\overrightarrow{h}^{e}_{i},\overleftarrow{h}^{e}_{i}]$. Then the context $cn_{t}$ is concatenated with the hidden state of the current decoder $\overrightarrow{h}^{d}_{t}$, and passed through a linear layer with the tanh activation function. Finally, predictions are generated by passing the activated output to another linear layer with a sigmoid activation function that generates a probability for each predicted item. The details of the model can be found in \citet{luong2015effective}.

\section{Instance Generation}
\label{Instance Generation}
We generate test instances with a varying number of periods up to 300 since lot-sizing has been utilized to generate daily solutions within a yearly planning horizon containing more than 360 periods \citep{darvish2016dynamic}. In addition, lot-sizing problems have been used with a large number of periods, such as 150 \citep{hartman2010dynamic}, 200 \citep{basnet2005inventory}, and 400 \citep{lu2011dynamic}, to test algorithmic advances. Similarly, knapsack problems with a large number of periods are used in practical applications, including advertising, container loading, airline ticketing, and processor scheduling \citep{papastavrou1996dynamic,hao2020dynamic}. We generate instances for both MCLSP and MSMK based on realistic data generation techniques \citep{buyuktahtakin2022stage} rather than using them from the literature since training requires many instances that such benchmark datasets do not contain.

\noindent \textbf{MCLSP Instances:} To generate instances, we employ the scheme presented in \citet{buyuktahtakin2018partial}. We have utilized MCLSP problems with varying sizes, characteristics, and horizons similar to those presented in the literature \citep{atamturk2004a, atamturk2005lot, hartman2010dynamic,buyuktahtakin2023scenario} resulting in a wide selection of computationally different problems to test our framework.  Two underlying parameters can simulate the problems of varying hardness levels: capacity-to-demand ratios $c \in \left\{10,14\right\}$ and setup-to-holding cost ratio $f=1,000$. The uniform integer distribution between $a$ and $b$ is denoted by $U\left[a,b\right]$. The shared capacity $c_{t}$ between items is sampled from $U\left[0.8c\bar{d},1.2c\bar{d}\right]$ where $\bar{d}$ is the overall average demand. The unit production cost $p_{it}$ is sampled from $U\left[1,200\right]$, the holding cost $h_{it}$ is sampled from $U\left[1,100\right]$, and the demand $d_{it}$ is sampled from $U\left[500,1500\right]$. The periodic setup cost is sampled from $U\left[0.9f\bar{h},1.1f\bar{h}\right]$ where $\bar{h}$ is the overall average holding cost. A planning horizon of $T=30$ with the number of items of $I=8$ and $T=40$ with $I=12$ is used in the training of two different encoder-decoder models. The first model with $I=8$ is used to predict instances with $T \in \left\{40,60,80,100,150,200\right\}$. In addition, the first model is used to predict item-wise generalization for instances with $I \in \left\{32,40,80\right\}$ and $T \in \left\{200,150,100\right\}$. The second model with $T=40$ with $I=12$ is used to predict instances with $T \in \left\{30,50,75,100,125,150\right\}$. Additionally, the second model is used to predict generalization instances in item order with $I \in \left\{36,48,60\right\}$ and $T \in \left\{125,100,80\right\}$. For MCLSP, a total of 18 sets of instances are used for testing, each containing 20 instances.

%

\noindent \textbf{MSMK Instances:} The instances are solved by sampling the problem parameters from the following integer uniform distributions. The profit $p_{it}$ of each item $i$ in period $t$ is sampled from $U\left[1,1000\right]$, the stability bonus $b_{it}$ from $U\left[1,1000\right]$, the weights of the items $w_{ijt}$ from $U\left[1,1000\right]$, and the capacity $c_{jt}$ from $U\left[0.5\sum_{i=1}^{I}w_{ijt},0.8\sum_{i=1}^{I}w_{ijt}\right]$. The number of training periods is $T =30$ and the number of items is $I \in \left\{8,10\right\}$. For the first model with $I=8$, testing instances with $T \in \left\{30,50,80,100,150,200\right\}$ are used. In addition, item-wise generalization test instances with $I \in \left\{16,24,32\right\}$ and $T \in \left\{30,20,15\right\}$ are used. For the second model with $I=10$, test instances with $T \in \left\{30,50,70,80,90,100\right\}$ and item-wise generalization test instances with $I \in \left\{20,30,40\right\}$ and $T \in \left\{30,20,15\right\}$ are used. For MSMK, 18 test sets are also generated in total, each having 20 instances.

\section{Model Training}
\label{Model Training Details}
In this section, we discuss the specific architectural details and hyperparameters of the trained models. In these models, the number of layers is two and the number of hidden units in the encoder is 128 for Table \ref{resultmclsp2}, 64 for Table \ref{resultmclsp3}, 256 for Table \ref{resultmsmk0}, and 128 for Table \ref{resultmsmk1}. The decoder for each model is twice the size of the encoder. Our choice of attention score mechanism is based on general score calculations. Our choice of optimizer for model training is the well-known Adam Optimizer \citep{kingma2014adam}, with an initial learning rate of 0.01 for Tables \ref{resultmclsp2} and \ref{resultmclsp3} and 0.001 for Tables \ref{resultmsmk0} and \ref{resultmsmk1}. We have used the dropout technique with a rate of $\left\{0.25,0.30,0.35\right\}$. It is a commonly used approach to prevent overfitting \citep{srivastava2014dropout}. We have also utilized a label smoothing approach to achieve better generalization \citep{muller2019does}. Our models are trained with 3,500,000 instances for each problem type. Training times are 18 and 30 hours for MCLSP with 8 and 12 items, respectively. Training times are 24 and 8 hours for MSMK with 8 and 10 items, respectively.

\section{PredOpt Reduces Objective Value Faster than CPLEX}
\label{PredOpt Reduces Objective Value Faster than CPLEX}
\begin{figure}[h!]
\centering
\subfloat[8 items with 60 periods\label{8item60period}]{%
  \includegraphics[width=0.49\textwidth]{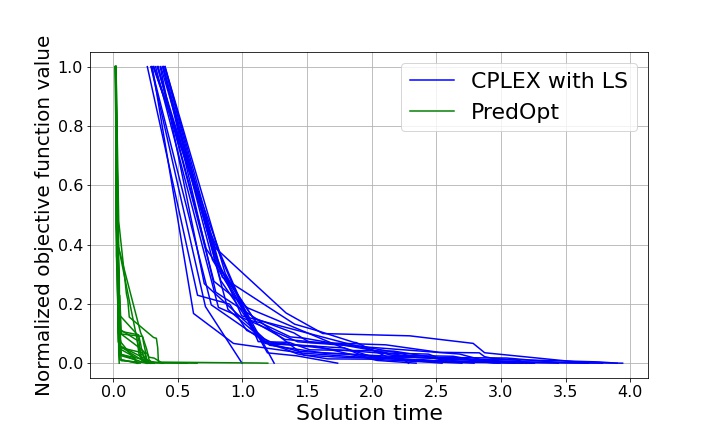}%
}%
\subfloat[8 items with 80 periods\label{8item80period}]{%
  \includegraphics[width=0.49\textwidth]{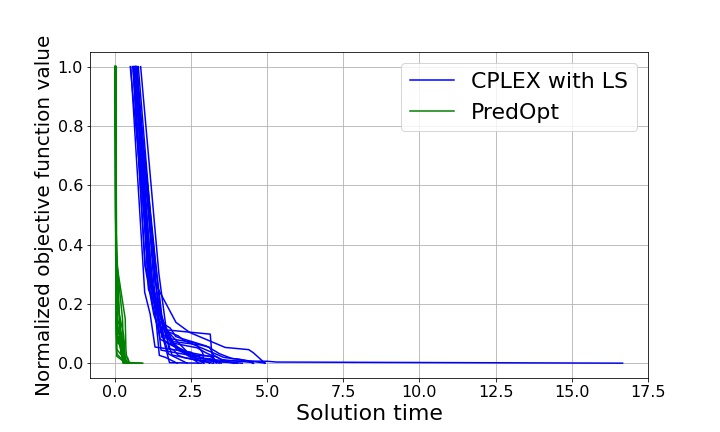}%
}

\subfloat[12 items with 50 periods\label{12item50period}]{%
  \includegraphics[width=0.49\textwidth]{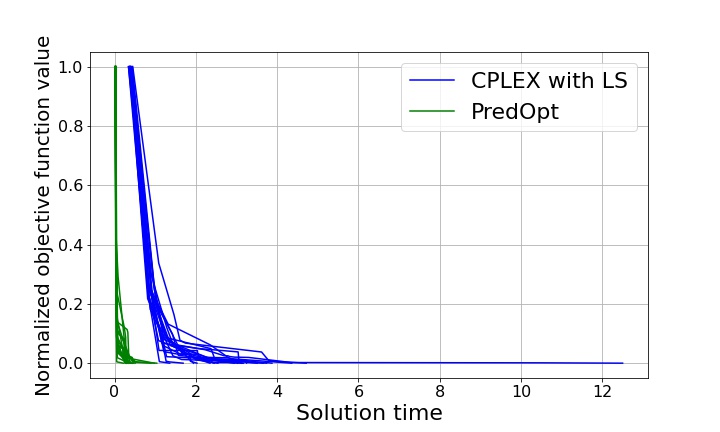}%
}%
\subfloat[12 items with 75 periods\label{12item75period}]{%
  \includegraphics[width=0.49\textwidth]{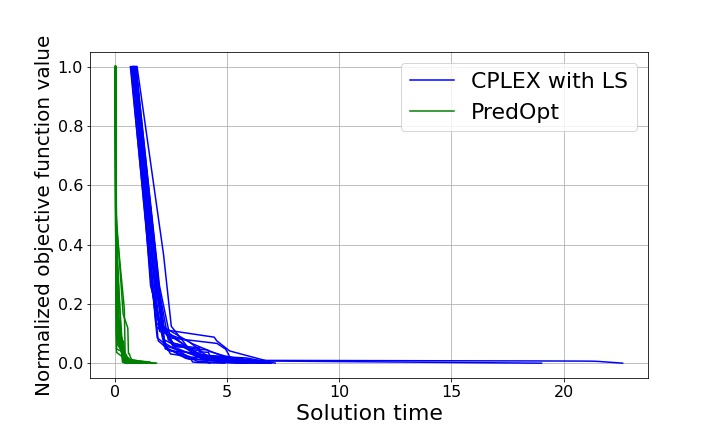}%
}
\captionsetup{justification=centering}
\caption{Progress of CPLEX with ($\ell$,S) inequalities and PredOpt objective values during the first few seconds of the solution process. All solution times are given in CPU seconds.}\label{figprogress}
\end{figure}

It is important to compare the speed of the PredOpt solution with CPLEX, since, in some cases, CPLEX quickly determines a good solution but takes a long time to prove the optimal one \citep{accorsi2022guidelines}. In Figure \ref{figprogress}, the progress during the solution process, in terms of a normalized objective function value averaged over 20 test instances, is presented to visualize the improvement of the PredOpt framework over CPLEX with ($\ell$,S) inequalities. Figures \ref{8item60period} and \ref{8item80period} present the progress of the second and third sets of test instances from Table \ref{resultmclsp2} with $T = 60,80$. Figures \ref{12item50period} and \ref{12item75period} present the progress for the second and third sets of test instances of Table \ref{resultmclsp3} with $T = 50,75$. As discussed in the previous results, Figure \ref{figprogress} demonstrates that the time of PredOpt is less than that of CPLEX with ($\ell$,S) inequalities to reach a particular objective function value. The graphs show that PredOpt reduces the objective function value much faster than CPLEX with ($\ell$,S). In all solution progress graphs, the objective values and solution times for PredOpt and CPLEX with ($\ell$,S) are visually quite distinguished, showing that the PredOpt framework helps improve the objective function quicker than CPLEX even in the first few seconds of the solution process.

\section{The Results of an Additional MSMK Generalization Experiment}
\label{appendixresultmsmkgen1}

Table \ref{resultmsmkgen1} presents another set of generalization experiments for MSMK. Here, we employ the trained model used in Table \ref{resultmsmk1} to generate results. The test instances contain up to $T=40$ and $I=200$. Similar to Table \ref{resultmsmkgen0}, the first three sets of instances have reduced the number of periods and increased the number of items. However, the last four sets of instances have both increased the number of items and periods compared to the data with which the model has trained. Also, none of these hard instances can be solved to optimality in the set 2-hour solution time limit. The results show that the PredOpt framework seems especially beneficial for such challenging instances compared to both heuristics. For example, in the seventh dataset with 200 items and 25 periods, PredOpt has an average solution time of 1.1 seconds with an optimality gap below 1\%, while the adaptive fixing heuristic and the problem reduction heuristic generate an inferior solution in a longer time. In addition, our PredOpt framework with the item-wise generalization algorithm can scale better to larger instances in terms of both solution time and quality compared to PR and AF heuristics.

\begin{table}[!htbp]
\caption{Average results of item-wise generalization experiments for the MSMK model trained with 10 items}
\label{resultmsmkgen1}
\centering
\begin{tabularx}{\textwidth}{rCCCCCCC}

Test Items	&	20	&	30	&	40	&	100	&	120	&	150	&	200	\\
$T$	&	30	& 	20	&  	15	&   	35	&    	40	&     	30 	&      	25	\\ \hline
timeCPX	&	3964.5	&	6225.0	&	4210.7	&	7211.0	&	7207.4	&	7210.8	&	7212.8	\\
timePredOpt	&	0.5	&	0.5	&	0.3	&	1.0	&	1.1	&	1.4	&	1.1	\\
timeAF	&	10.2	&	6.7	&	5.1	&	29.7	&	43.9	&	34.6	&	31.1	\\
timePR	&	60.0	&	70.2	&	15.9	&	526.5	&	572.5	&	471.1	&	487.9	\\
timeImpPredOpt	&	9029	&	13268	&	13536	&	6987	&	6540	&	5353	&	6763	\\
timeImpAF	&	372	&	955	&	874	&	246	&	166	&	210	&	236	\\
timeImpPR	&	734	&	2206	&	2316	&	74	&	14	&	116	&	230	\\
p-value	&	$ \leq 0.001$	&	$ \leq 0.001$	&	$ \leq 0.001$	&	$ \leq 0.001$	&	$ \leq 0.001$	&	$ \leq 0.001$	&	$ \leq 0.001$	\\
accuracy(\%)	&	94.0	&	93.2	&	90.5	&	89.4	&	87.9	&	93.0	&	92.8	\\
optGapPredOpt(\%)	&	0.66	&	0.67	&	1.77	&	1.81	&	2.69	&	0.79	&	0.91	\\
optGapAF(\%)	&	1.50	&	1.25	&	1.02	&	3.83	&	3.65	&	3.67	&	3.59	\\
optGapPR(\%)	&	2.28	&	4.37	&	6.31	&	3.18	&	3.75	&	4.96	&	5.55	\\ \hline

\end{tabularx}
\end{table}







\end{document}